\documentclass{osa-article}

\journal{oe}

\articletype{Research Article}

\usepackage{lineno}

\usepackage[ruled,vlined,linesnumbered]{algorithm2e}
\usepackage{mathrsfs}
\usepackage{xcolor}

\definecolor{darkblue}{rgb}{0.0, 0.0, 0.55}
\definecolor{darkcerulean}{rgb}{0.03, 0.27, 0.49}

\newcommand\change[1]{{\color{black}#1}}

\newcommand\tocheck[1]{{\color{black}#1}}

\newcommand\Mat[1]{{\mathbf{#1}}}
\newcommand\Vect[1]{{\mathbf{#1}}}
\DeclareMathOperator*{\argmin}{arg\,min}
\newcommand\psf{{\mathrm{\mathbf{PSF}}}}
\newcommand\mtf{{\mathrm{\mathbf{MTF}}}}

\newcommand\mtfv{{\mathrm{\mathbf{MTFv}}}}
\newcommand\Fr[1]{{\mathscr{F}_z\left\{#1\right\}}}
\newcommand\ft[1]{{\mathcal{F}\left\{#1\right\}}}

\begin{document}

\title{Diffractive lensless imaging with optimized Voronoi-Fresnel phase}

\author{Qiang Fu,\authormark{1,*} Dong-Ming Yan,\authormark{2,3} and Wolfgang Heidrich\authormark{1}}

\address{\authormark{1}King Abdullah University of Science and Technology (KAUST), Thuwal 23955-6900, Saudi Arabia\\
\authormark{2}National Laboratory of Pattern Recognition (NLPR), Institute of Automation, Chinese Academy of Sciences, 100190, Beijing, China\\
\authormark{3}School of AI, University of Chinese Academy of Sciences, 101408, Beijing, China}

\email{\authormark{*}qiang.fu@kaust.edu.sa} 



\begin{abstract}
Lensless cameras are a class of imaging devices that shrink the physical dimensions to the very close vicinity of the image sensor by replacing conventional compound lenses with integrated flat optics and computational algorithms. Here we report a diffractive lensless camera with spatially-coded Voronoi-Fresnel phase to achieve superior image quality. We propose a design principle of maximizing the acquired information in optics to facilitate the computational reconstruction. By introducing an easy-to-optimize Fourier domain metric, Modulation Transfer Function volume (MTFv), which is related to the Strehl ratio, we devise an optimization framework to guide the optimization of the diffractive optical element. The resulting Voronoi-Fresnel phase features an irregular array of quasi-Centroidal Voronoi cells containing a base first-order Fresnel phase function. We demonstrate and verify the imaging performance for photography applications with a prototype Voronoi-Fresnel lensless camera on a 1.6-megapixel image sensor in various illumination conditions. Results show that the proposed design outperforms existing lensless cameras, and could benefit the development of compact imaging systems that work in extreme physical conditions.
\end{abstract}

\section{Introduction}
Imaging devices, such as photographic objectives and microscopes, have long been relying on lenses to focus light and create a projection of the scene onto photosensitive sensors. In such designs, as in human eyes, the focal length of the lens presents a fundamental limit for the overall device form factor. In addition, various optical aberrations preclude the use of single lenses for wide field-of-view (FOV) imaging. Instead, groups of compound lenses that are carefully designed must be used for good image quality.

With the ever increasing demand for compactness of imaging optics, great efforts have been motivated to develop lensless cameras during the past few years. To date two prevailing lensless methods exist, natural compound-eye mimicry~\cite{wu2017artificial} and heuristic point spread function (PSF) engineering~\cite{boominathan2022recent}. 

Analogous to insect eyes, various artificial compound-eye designs have been proposed to directly mimic the eye structures in nature. Early artificial compound-eye structures tessellate regular Micro-Lens Arrays (MLAs) on planar or curved surfaces. Each lenslet is an independent unit to create a small image. By tessellating an array of lenslets on a planar surface, TOMBO~\cite{tanida2001thin,horisaki2010irregular} produces final images by a backprojection algorithm to stitch sub-images together. To prevent optical cross-talks, a light-blocking layer has to be employed under each micronlens. The number of output image pixels is equal to the number of lenslets. Gabor superlens~\cite{stollberg2009gabor} employs sophisticated multi-layer planar MLAs and aperture arrays along the optical path to re-arrange light rays on the sensor. Followup implementations either stitch the sub-images algorithmically (eCley~\cite{bruckner2010thin}) or optically (oCley~\cite{meyer2011optical}). A recent artificial ommatidia lensless camera~\cite{kogos2020plasmonic} employs plasmonic structures to allow for larger FOV for planar compound eyes. Tessellating lenslets on curved substrates is more challenging, but yields a wide FOV. A digital compound-eye camera~\cite{song2013digital} tessellates a uniform elastomeric micro-lens array with stretchable electronics on a hemisphere to fully resemble the arthropod eyes. CurvACE~\cite{floreano2013miniature} achieves 180$^{\circ}$ horizontal FOV and 60$^{\circ}$ vertical FOV with polymer microlenses and flexible printed circuits. These methods focus more on the optical tessellation, with less attention on the computational reconstruction. The resolution is relatively low, and only simple scene targets can be imaged.

The other strategy considers the entire system from the perspective of the PSF it generates. Instead of mapping each scene point to a single focus spot on the image plane, as in conventional lens-based systems, such lensless cameras render each scene point as a distributed pattern that covers large areas of the image plane. The captured raw data is interpreted by computational imaging algorithms based on the physical model to reconstruct the latent image.  Early implementations make use of amplitude masks~\cite{boominathan2016lensless} to create patterns induced by shadowing effects. In order to improve the numerical conditioning, separable masks have been proposed in a coded-aperture design~\cite{deweert2015lensless} and FlatCam~\cite{asif2016flatcam} for optimal amplitude mask design. Fresnel Zone Aperture (FZA)~\cite{wu2020single} is also used for lensless imaging. These amplitude masks inherently suffer from low light efficiency. Phase-only lensless cameras improve the overall throughput without blocking light. A binary phase profile with tailored odd-symmetry gratings in PicoCam~\cite{gill2013odd,stork2013lensless} produces spiral-shaped PSFs. An important pioneer work is DiffuserCam~\cite{antipa2018diffusercam}, where a non-optimized diffuser is used to generate caustic patterns as the PSF. A Perlin pattern is later introduced in PhlatCam~\cite{boominathan2020phlatcam} as a better heuristic PSF. Random lenslets have also been favored in 3D light field microscopy, such as Miniscope3D~\cite{yanny2020miniscope3d} and Fourier diffuserScope~\cite{liu2020fourier}. A random lenslet diffuser has also been used as a lensless on-chip fluorescence microscope~\cite{kuo2020chip}. \change{A custom microlens array (MLA) has been adopted as a single optical component in the computational miniature mesoscope~\cite{xue2020single,xue2022deep} for fluorescence imaging. Very recently a learned 3D lensless camera~\cite{tian2022learned} co-designs the MLA and the neural network to achieve single-shot 3D imaging without the need of PSF calibration.}

Here we report a new diffractive lensless camera with spatially-coded Voronoi-Fresnel phase to improve the imaging performance by leveraging the benefits of both the compound-eye structure and PSF engineering. Inspired by the random and uniform distribution of lenslets in the apposition compound eyes~\cite{kim2016hexagonal}, and motivated by an observation from the properties of the engineered heuristic PSFs~\cite{boominathan2020phlatcam}, we find that the image quality after algorithmic reconstruction is positively related to the sparse distribution of high-contrast bright spots in the PSFs, subject to the phase being physically realizable. In other words, lensless cameras favor PSFs with concentrated sparse patterns. From the perspective of Fourier optics, Modulation Transfer Function (MTF)~\cite{goodman2005introduction}, the Fourier counterpart of PSF, offers a more comprehensive measure to quantify this phenomenon. High frequency details are better recoverable in the reconstruction algorithms if the cut-off frequency is kept as large as possible, as well as the MTF is uniformly distributed in all directions in the Fourier domain. However, MTF is a 2D function of spatial frequencies. We therefore define MTF volume (MTFv) as a single-number metric for information maximization. A larger MTFv value encourages the system to transmit more information from hardware to software. 

By applying this principle, we find that a certain number of individual point PSFs makes an excellent match to the preferred PSFs in lensless imaging. To this end it is possible to engineer the optimal PSFs from the compound-eye mimicry. We devise a metric-guided optimization framework with modified Lloyd's iterations to find non-trivial and non-heuristic solutions to the problem. The resulting PSF is a collection of diffraction limited diverse directional spots, with optimized spatial locations and total number. They do not work individually as in conventional compound eyes, but their union is the equivalent PSF to be coupled with computational algorithms to reconstruct sharp images. With significant lower requirement in optics and more flexibility in computation, the proposed method offers a solution to high quality 2D photography devices that work in extreme physical conditions, such as wearable cameras, in-cabin monitoring, and capsule endoscopy. Unconventional imaging applications in various disciplines are expected to be triggered owing to the emerging importance of flat optics imaging.


\section{Methods}
\subsection{Overview}
\begin{figure*}[h!]
    \centering
    \includegraphics[width=\textwidth]{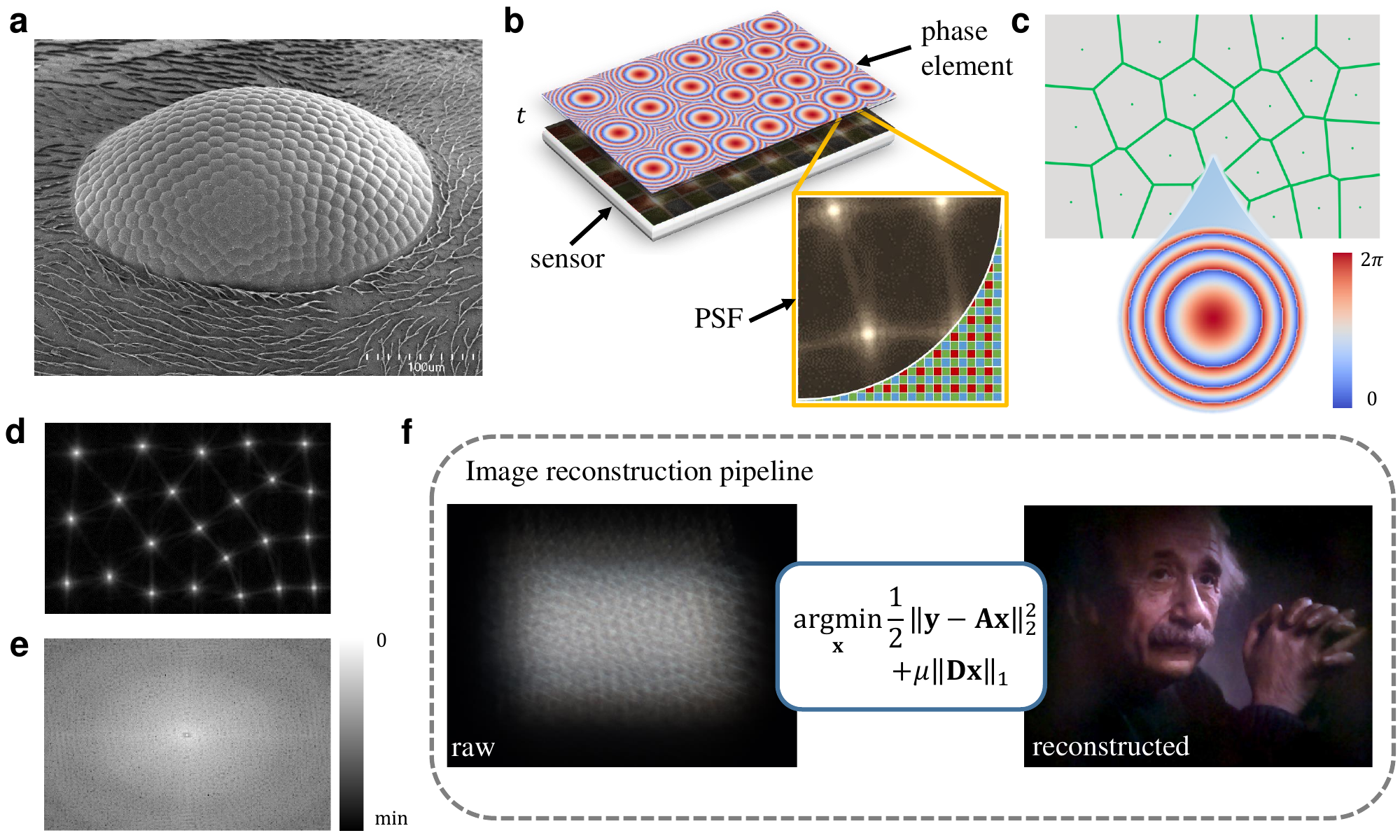}
    \caption{Overview of Voronoi-Fresnel lensless imaging. (a) Micrograph of the compound eyes of an \textit{Iridomyrmex purpureus} (image courtesy of Roberto Keller, ANTWEB1008536, from www.antweb.org). (b) The proposed Voronoi-Fresnel lensless camera consists only of a phase element in close proximity to the sensor (distance $t$ is a few $mm$). The zoom-in illustrates the detailed PSF and Bayer filters. (c) The first-order Fresnel phase function is the building block of the camera. Their center locations are distributed in a quasi-Centroidal Voronoi Tessellation across the 2D plane. (d) The PSF is an array of diffraction limited spots with optimal spatial locations (intensity enhanced for better visualization). (e) MTF (in log scale) is uniform across the Fourier domain. (f) The image reconstruction pipeline converts the uninterpretable raw data (left) to a high quality image (right).}
    \label{fig:overview}
\end{figure*}

An overview of the proposed Voronoi-Fresnel lensless camera is shown in Fig.~\ref{fig:overview}. Inspired from the delicate distributions of the ant's ommatidia (Fig.~\ref{fig:overview}a), our lensless camera consists of an optimized phase element just a few millimeters above the image sensor (Fig.~\ref{fig:overview}b). The phase features an irregular array of quasi-Centroidal Voronoi cells containing a base first-order Fresnel phase function (Fig.~\ref{fig:overview}c). By applying the proposed design principle, our Voronoi-Fresnel phase yields optimized PSF (Fig.~\ref{fig:overview}d) and MTF (Fig.~\ref{fig:overview}e). An image reconstruction pipeline (Fig.~\ref{fig:overview}f) is then employed to recover high-quality images from the non-interpretable raw data.

\subsection{Voronoi-Fresnel phase}
Our Voronoi-Fresnel phase is composed of a base first-order Fresnel phase that is duplicated in various sub-regions on the 2D plane (Fig.~\ref{fig:overview}c). The base Fresnel phase at the design wavelength $\lambda$ is defined as 
\begin{equation}
\Phi_F \left(\xi, \eta, \lambda \right) = \exp{\left[ -j k \left( {\xi}^2 + {\eta}^2 \right) / \left(2z\right) \right]},
\end{equation}
where $\left(\xi, \eta\right)$ are the Cartesian coordinates on the phase plane, $k = 2 \pi / \lambda$ is the wave number, and $z$ is the distance from the optics to the image sensor. This phase function is the first-order approximation of an ideal lens, producing a diffraction limited spot in the paraxial region.

We divide the design space into regions that form a complete tessellation of the whole area. A typical tessellation is a Voronoi diagram, which is a collection of sub-regions that contain points closer to the corresponding generating sites than any other sites. Each sub-region, also known as a Voronoi cell $V_i$, features a center location and a few vertices. The origin of the Fresnel phase function coincides with the center location, and the polygon determined by the vertices creates a distinct aperture for that cell. We refer to each Voronoi cell with the Fresnel phase as a Voronoi-Fresnel cell. The entire Voronoi-Fresnel phase is a collection of all the constituent Voronoi-Fresnel cells,
\begin{equation}
\Phi \left(\xi, \eta, \lambda \right) = \sum_{i=1}^{K} A_{i} \left(\xi - \xi_{i}, \eta - \eta_{i} \right) \cdot \exp{\left( -j \frac{2\pi}{\lambda} \cdot \frac{\left(\xi - \xi_{i}\right)^2 + \left(\eta - \eta_{i}\right)^2}{2z}\right)},
\label{eq:fresnelphase}
\end{equation}
where $\left(\xi_{i}, \eta_{i}\right)$ are the center locations of all the $K$ Voronoi-Fresnel cells. The aperture function $A_{i}$ is defined by the vertices of the $i$-th sub-region, where $A_{i} = 1$ inside $V_{i}$, and $A_{i} = 0$ outside. The union of all cells is the whole region, and any two different cells have no intersections. 

\subsection{PSF and the MTFv metric}
The PSF is characterized by the Fresnel diffraction pattern of the entire Voronoi-Fresnel phase on the image sensor, which is the squared magnitude of the diffracted optical field. The panchromatic PSF can then be calculated as an integral of spectral PSFs over the effective spectral range  $[\lambda_1, \lambda_2]$, 
\begin{equation}
\psf \left(x, y\right) = \int_{\lambda_1}^{\lambda_2} \left| \Fr{\Phi \left(\xi, \eta, \lambda \right)} \right|^2 \mathrm{d} \lambda,
\label{eq:panpsf}
\end{equation}
where $\Fr{\cdot}$ is the Fresnel propagator for a distance $z$. 

Since the entire phase function is a collection of the base Fresnel phase, we further investigate the individual PSFs that are generated from the polygonal sub-apertures. As demonstrated in \tocheck{Supplemental Document 1}, when adjacent Fresnel phase centers are at the centroids of the cell, the cross interference between individual cells are negligible, with maximum error less than 1\%. The equivalent PSF can be approximated by the union of individual PSFs,
\begin{equation}
\psf \left(x, y, \lambda\right) \approx  \sum_{i=1}^{K} \psf_i^0 \left(x - \xi_i, y - \eta_i, \lambda\right),
\end{equation}
where $\psf_i^0$ is the centered PSF as if the aperture were located at the origin. Note that in the simulation and optimization below, we do not simulate individual PSFs separately, but use the entire constructed phase function to obtain the panchromatic PSF.

\begin{figure*}[htp]
    \centering
    \includegraphics[width=\textwidth]{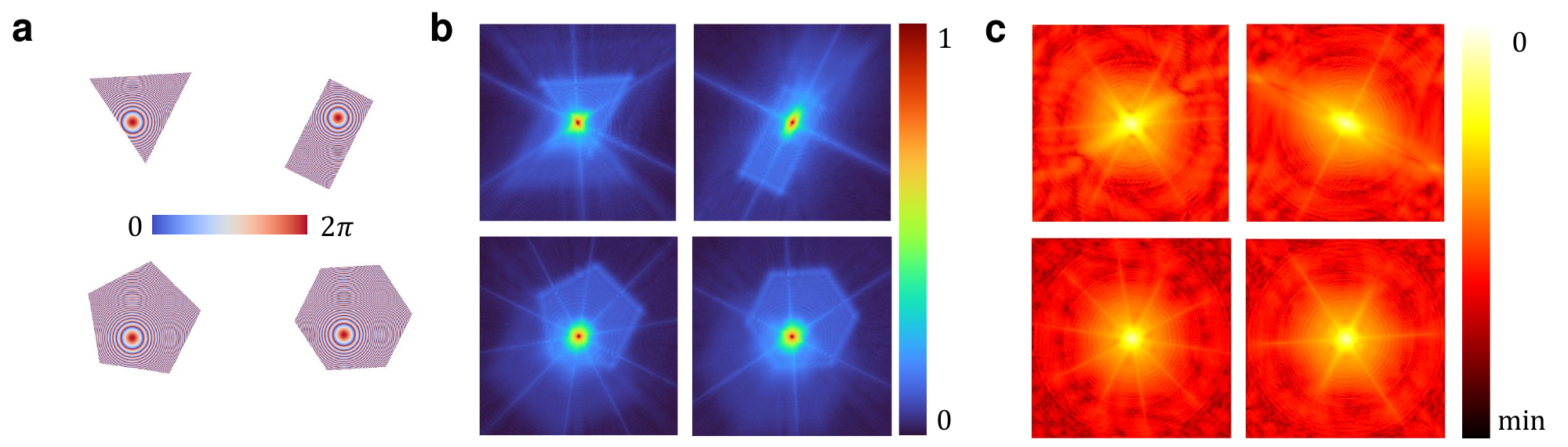}
    \caption{Directional filtering effects. (a) Base Fresnel phase with various aperture geometries. Top: triangle and \change{rectangle}. Bottom: pentagon and hexagon. (b) The  \change{panchromatic PSFs (brightness enhanced) are shown in false color}. (c) The corresponding MTFs (in log scale) show distinct directional filtering effects.} 
    \label{fig:directional}
\end{figure*}

These irregular apertures are significant, since the resulting diffraction pattern generates a set of compact directional filters depending on the geometries. We show such directional filtering effect in Fig.~\ref{fig:directional}. The center of a base Fresnel phase is first shifted randomly off the origin. A polygon aperture is then imposed on the phase profile. Here, without loss of generality, we illustrate four simple polygons, a triangle, a \change{rectangle}, a pentagon, and a hexagon (Fig.~\ref{fig:directional}a). The corresponding \change{panchromatic} PSFs are rendered \change{in false color} in Fig.~\ref{fig:directional}b. Long tails in the PSFs show up in the perpendicular directions to the polygon edges due to diffraction. A collection of such directional spots resulted from these cells make up the effective PSF.

It is more effective to evaluate the PSF through its frequency counterpart, the MTF. With higher MTF across all frequencies, the imaging system preserves more details in the scene, a property that has long been used to evaluate the quality of lens-based optics. It can be seen from the log-scale panchromatic MTFs in Fig.~\ref{fig:directional}c that the long tails in the spatial domain cause the MTF to drop in certain directions, while information is maintained in other directions. \change{For example, in the triangle and rectangle cases, the elongated directions in the PSF are compressed in the MTF, whereas the squeezed directions in the PSF are then pulled in the frequency domain. This is a result of the scaling property of the Fourier transform.} As the polygon becomes more regular in all edges \change{(e.g., the pentagon and hexagon examples)}, the directional filtering tends to be more isotropic. A circular aperture as in conventional lenses would result in no directional filtering effect at all.

The distribution of the panchromatic PSF, or equivalently the MTF, is key to the imaging performance. However, MTF is a 2D function in the Fourier domain. It is difficult to use directly as a metric to optimize the optical element. We propose to use normalized panchromatic MTF volume (MTFv) as a single-number metric to evaluate the system performance,
\begin{equation}
\mtfv = \frac{1}{\left| \Omega \right|} \iint \mtf \left(f_X, f_Y\right) \mathrm{d} f_X \mathrm{d} f_Y ,
\label{eq:mtfv}
\end{equation}
where $f_X$ and $f_Y$ are the Fourier frequencies, and $\left| \Omega \right|$ is the area of the design space. The MTF is the Fourier transform of the panchromatic PSF, i.e., $\mtf \left(f_X, f_Y,\right) = \left| \ft{ \psf \left(x, y\right) } \right|$, with $\ft{\cdot}$ being the Fourier transform. 

The MTFv is not only simple for computation, but also shares a connection with the well-established optics metric, Strehl ratio, that is widely used in image quality evaluation. As derived in \tocheck{Supplemental Document 1}, Strehl ratio can be approximated by integrating the MTF as an upper bound~\cite{roberts2002characterization}. Since the reference diffraction limited quantity only affects the denominator, we can ignore it when used as a figure-of-merit, leading to the above MTFv metric. In addition, MTFv is in principle a generic metric that imposes no restrictions on the intensity distributions of the PSFs, so it is applicable for all phase-type lensless designs. A larger MTFv value encourages more information to be recorded by the optical system, so we can employ it as a guide to seek for optimal PSFs generated by the Voronoi-Fresnel phase.

\subsection{Optimization and properties}
MTFv is a highly nonlinear and non-convex function of the number of Voronoi regions $K$ and their center locations $\left(\xi_i, \eta_i\right)$. It is challenging to optimize the MTFv over the entire parameter space. Therefore, we do not aim at finding a closed-form solution to maximizing MTFv. Instead, we devise an optimization framework to search for a feasible solution. We show in the \tocheck{Supplemental Document 1} that the effective MTF largely depends on three factors, the diffraction by each aperture; the phase delay terms in the Fourier domain that are introduced by the amount of spatial shifts from the centered PSFs; and the total number of the Voronoi-Fresnel cells $K$. 

The smallest dimension in the aperture geometry plays an important role due to diffraction. The phase delay terms would vary dramatically as the distances between each pair of center locations change. Given a certain number of sites $K$ within a fixed area, there are infinite Voronoi tessellations, whereas the MTFv values vary significantly. Taking the extreme case of $K = 1$, we get a compact single spot PSF, but it is impractical to realize such an optical element that covers the whole image sensor. On the other hand, when $K$ approaches infinity, the PSF becomes pixel-wise uniform, and carries no useful information. Therefore, there exists an optimal number $K$ that maximizes the MTFv.

Our solution to the Voronoi-Fresnel phase optimization is based on the above conjecture that the optimal center locations would require uniform distributions of the Voronoi cells over the 2D design space, as well as keeping some degrees of irregularity of each Voronoi region to diversify the spatial filtering of individual PSFs. To take the above factors into account, our optimization framework employs a two-step search scheme. In the first step, we fix $K$, and adopt a quasi-CVT routine to maximize the panchromatic MTFv, as summarized in \tocheck{Algorithm~\ref{alg:max_mtfv}}. The second step is then a parameter sweep to find the best $K$.

{\centering
\begin{minipage}{0.85\textwidth}
\begin{algorithm}[H]
\SetAlgoLined
\KwIn{number of Voronoi regions $K$, fabrication parameters; \textit{maxiter}, \textit{tol}}
\KwOut{Voronoi sites \textit{coords}, \textit{maxMTFv}}
\BlankLine
generate $K$ random points $\left\{\Vect{p}_i\right\}_{i=1}^{K}$\;
construct Voronoi tessellation $\left\{V_i\right\}_{i=1}^{K}$\;
compute MTFv by Eq.~\eqref{eq:mtfv}\;
\textit{coords} $\longleftarrow \left\{\Vect{p}_i\right\}_{i=1}^{K}$\;
\textit{maxMTFv} $\longleftarrow$ \textit{MTFv}\;
\For{k $\leq$ maxiter}{
  compute mass centroids $\Vect{c}_i$ for $\left\{V_i\right\}_{i=1}^{K}$\;
  update $\left\{\Vect{p}_i\right\}_{i=1}^{K} \longleftarrow \left\{\Vect{c}_i\right\}_{i=1}^{K}$\;
  construct new Voronoi tessellation\;  
  compute new \textit{MTFv}\;
  \eIf{MTFv $>$ maxMTFv}{
   \textit{coords} $\longleftarrow \left\{\Vect{p}_i\right\}_{i=1}^{K}$\;
   \textit{maxMTFv} $\longleftarrow$ \textit{MTFv}\;
   }{
   continue\;
  }
  \textit{error} = $\sqrt{\frac{1}{K} \sum_{i=1}^{K} \left\Vert\Vect{p}_i^{k} - \Vect{p}_i^{k-1}\right\Vert^2}$\;
  \If{error $<$ tol}{
  break\;
  }
}
\caption{Maximize MTFv for a given $K$}
\label{alg:max_mtfv}
\end{algorithm}
\end{minipage}
\par
}

In the first step, we initialize a set of $K$ random coordinates. A Voronoi tessellation is constructed to produce the Voronoi-Fresnel phase function by Eq.~\eqref{eq:fresnelphase}. The corresponding MTFv is computed by Eq.~\eqref{eq:mtfv}. In each iteration, we compute the centroids of each cell and update the center coordinates with the centroids. A new Voronoi tessellation is constructed, leading to a new MTFv for the updated Voronoi-Fresnel phase. The new site coordinates and the corresponding MTFv are recorded only if the current MTFv is larger than the existing maximum MTFv, otherwise are discarded. After each iteration, we compute the root-mean-square (RMS) distance between the new sites and the previous sites as the residual error. The iterations terminate when the residual error is smaller than a preset tolerance. To make the algorithm fabrication-aware, we set the terminating tolerance \textit{tol} as a hundredth (0.01) of the smallest feature size that can be fabricated. The iterations terminate after a maximum number of iterations \textit{maxiter} is reached.

\begin{figure*}[h!]
    \centering
    \includegraphics[width=\textwidth]{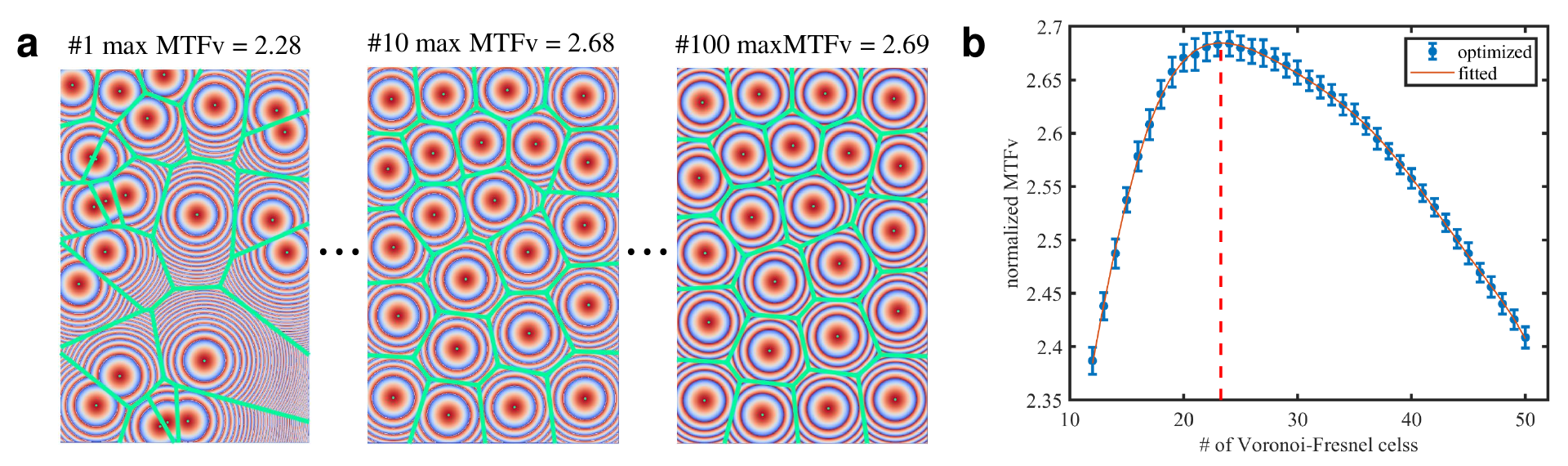}
    \caption{Optimization of Voronoi-Fresnel phase. (a) Given a fixed number of Voronoi-Fresnel cells $K=23$, as the optimization evolves (see \tocheck{Visualization 1}), the initial (left, Iteration \#1) random distribution of Voronoi-Fresnel cells gradually tends to be more uniform (middle, Iteration \#10) while keeping individual irregular apertures. The corresponding MTFv values increase until convergence (right, Iteration \#100). (b) After the second step of parameter sweep on $K$, a plot of MTFv with respect to $K$ can be fitted to a polynomial to find the optimal number of $K = 23$ in this example. \change{The MTFv values for each $K$ are the mean of 100 runs with different random initialization, and the error bar is the corresponding standard deviations.}} 
    \label{fig:optimization}
\end{figure*}

An exemplary quasi-CVT optimization is shown in Fig.~\ref{fig:optimization}a followed by a parameter sweep step in Fig.~\ref{fig:optimization}b. This simulation uses 240~$\times$~160 sensor pixels with a pixel pitch of 3.45~$\mu$m. The distance is 2~mm from the optical plane to the sensor plane. The quasi-CVT optimization is initialized by a random set of points as the center locations of the base Fresnel phase (green dots). The green edges mark the individual apertures. As the optimization evolves, the Voronoi-Fresnel cells tend to scatter uniformly in the design space, while a certain degree of irregularity of the apertures is maintained. When the optimization converges, an optimal phase profile is achieved. See \tocheck{Visualization 1} for an animation of the optimization process. \change{To account for the random initialization effects, for each $K$ we run the quasi-CVT steps 100 times with different initializations, and take the mean value as the resulting MTFv. We take the standard deviation among the 100 runs as the uncertainty (Fig.~\ref{fig:optimization}b). After repeating optimization for different $K$ values in the same manner, the mean MTFv can be plotted against $K$, and a polynomial curve is fitted to the data.} The maximum MTFv is found to occur when $K = 23$ in this example. The corresponding Voronoi-Fresnel phase is selected as the final result.

\begin{figure*}[h!]
    \centering
    \includegraphics[width=\textwidth]{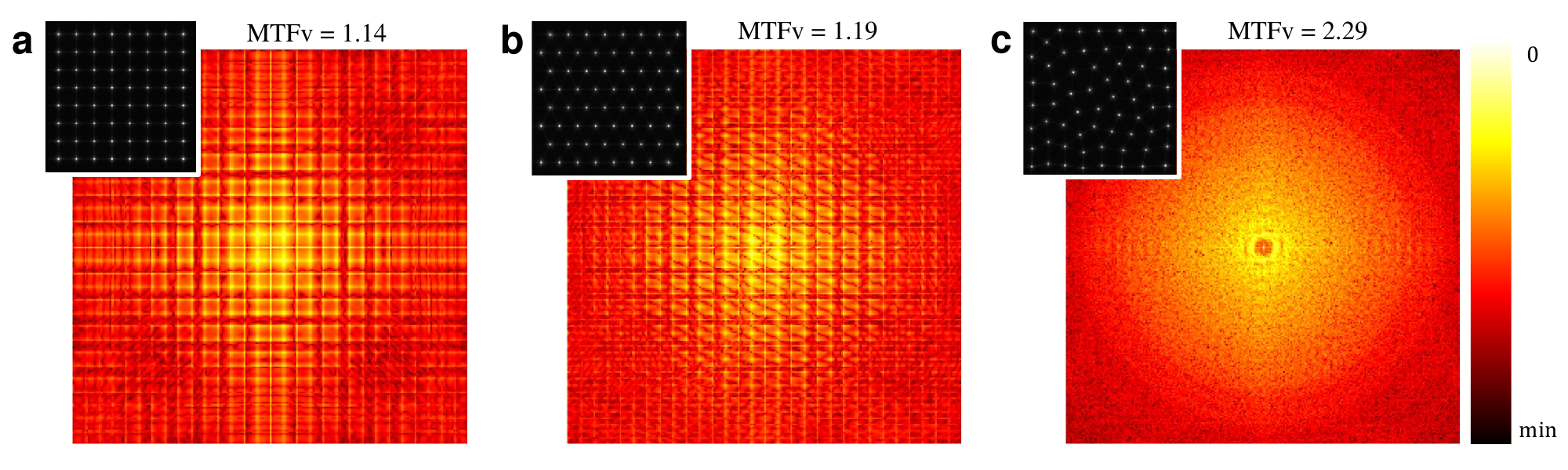}
    \caption{MTF comparison. (a) Rectangular grid. (b) Hexagonal grid. (c) Voronoi-Fresnel. The optimized quasi-CVT shows more uniform frequency distribution and higher MTFv values. All MTFs are shown in log scale. The corresponding PSFs are shown in the top-left corners in each case.} 
    \label{fig:properties}
\end{figure*}

A notable difference between our quasi-CVT algorithm and conventional CVT is that, a certain degree of irregularity should be preserved in quasi-CVT, while the CVT tends to converge to a regular hexagonal grid. Theoretically, the optimal CVT in a 2D plane results in a hexagonal pattern~\cite{yan2011computing}. To demonstrate that the optimization is non-trivial, we compare the optimized result from the above algorithm against a regular rectangular and a hexagonal Voronoi-Fresnel phase in Fig.~\ref{fig:properties}. The design area is 256~$\times$~256 pixels, and the number of cells is 64 in all three cases. The two regular grids are equivalent to common off-the-shelf microlens arrays. The MTFv on the regular rectangular (MTFv = 1.14) or hexagonal grid (MTFv = 1.19) is significantly lower than the optimized result (MTFv = 2.29). Periodic patterns exist in both regular grids. This attributes to the fact that the regular grids lack of diversity in the directional filtering. Therefore, the optimization is non-trivial. Simple and regular off-the-shelf microlens arrays are not well-suited for this purpose.

The parameter sweep in the second step of the algorithm is time-consuming, especially when the design space is large. We can facilitate the search of best number of sites in the full scale from smaller scales. We evaluate the optimal $K$ in various common scales for various aspect ratios of 1:1, 4:3, 3:2, 16:9, 21:9, and plot the MTFv with the design area in Fig.~\ref{fig:scalability}. See detailed analysis in \tocheck{Supplemental Document 1}. The result indicates our algorithm is linearly scalable with respect to the design area, regardless of the aspect ratios of the design area.

\begin{figure}[h!]
    \centering
    \includegraphics[width=0.6\columnwidth]{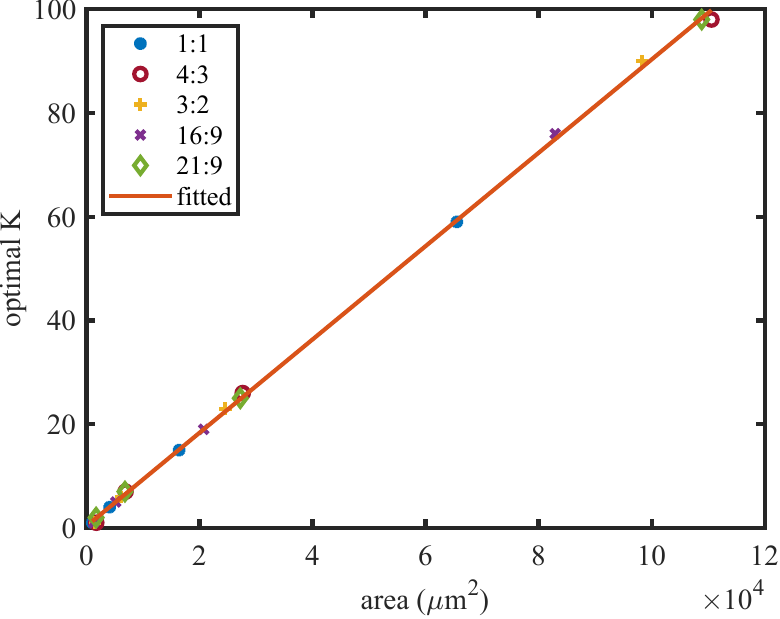}
    \caption{Linear Scalability. The optimal number of Voronoi sites is linearly proportional to the design area, regardless of the aspect ratios of the design areas}.
    \label{fig:scalability}
\end{figure}

\subsection{Image reconstruction}
We formulate the image formation process as a convolution between the ground-truth image and the PSF. It can be expressed in the matrix-vector product form, 
\begin{equation}
\Vect{y} = \Mat{A} \Vect{x} + \Vect{n},  
\end{equation}
where $\Vect{x} \in \mathbb{R}^{3N}$ is the ground-truth image with $N$ pixels in three color channels, $\Mat{A} \in \mathbb{R}^{3N \times 3N}$ is a matrix that represents the color PSF, and $\Vect{y} \in \mathbb{R}^{3N}$ is the captured raw data. The data is degraded by additive noise $\Vect{n}$.

There are various methods to solve the image based on the above image formation model. A straightforward way is to solve an inverse problem with effective image priors in an optimization framework. Recent data-driven image reconstruction methods make use of  end-to-end deep neural networks to inference the required image. They usually require a large dataset captured with specific designs, such as diffuserCam~\cite{monakhova2019learned,bae2020lensless}, or PhlatCam\cite{khan2020flatnet}, and cannot be generalized from hardware to hardware. In this paper, we focus on the advantages of the optical optimization, so we adopt the image deconvolution method with a Total Variation (TV) regularization to demonstrate and compare the optical advantages agianst prior works. The TV regularizer encourages the sparsity of edges in natural images, which has proved to be effective in lensless image reconstruction~\cite{antipa2018diffusercam,boominathan2016lensless}. Specifically, we solve
\begin{equation}
\argmin_{\Vect{x}} \frac{1}{2} \left\Vert \Mat{A} \Vect{x} - \Vect{y} \right\Vert^2_2 + \mu \left\Vert \Mat{D} \Vect{x} \right\Vert_{1},
\end{equation}
where $\left\Vert \cdot \right\Vert_2$ and $\left\Vert \cdot \right\Vert_1$ are the $\ell_2$-norm and $\ell_1$-norm respectively. $\Mat{D}$ is the finite difference operator, and $\mu$ is a penalization weight. An effective way to solve this problem is to employ the Alternating Direction Method of Multipliers (ADMM)~\cite{boyd2011distributed}. The detailed implementation is presented in \tocheck{Supplemental Document 1}.

\section{Results}
\subsection{Simulation results}
We first validate our methods in simulation, and compare its performance with very recent phase-type lensless cameras that share the most similar inspirations, DiffuserCam~\cite{antipa2018diffusercam} and PhlatCam~\cite{boominathan2020phlatcam}. To ensure a fair comparison, all the simulations are performed with the same conditions. We assume the sensor pixel pitch is 4~$\mu$m, and the resolution of the phase element is 1~$\mu$m, which provides a good sampling rate between the two. The sensor pixels are 256~$\times$~256, and Voronoi-Fresnel phase is 1024~$\times$~1024. The distance between the optical element and the sensor is 2~mm. The PSFs for the Voronoi-Fresnel design and the PhlatCam design are obtained through a full spectrum simulation, from 400~nm to 700~nm with an interval of 10~nm (totally 31 spectral bands). We are not able to simulate a diffuser PSF, so we adopt the closest PSF from DiffuserCam, and assume the intensities are the same for the three color channels. To take the spectral variations into account, we use multispectral image data~\cite{yasuma2010generalized,li2021multispectral} for the full spectrum simulation. 

\begin{figure*}[h!]
    \centering
    \includegraphics[width=\textwidth]{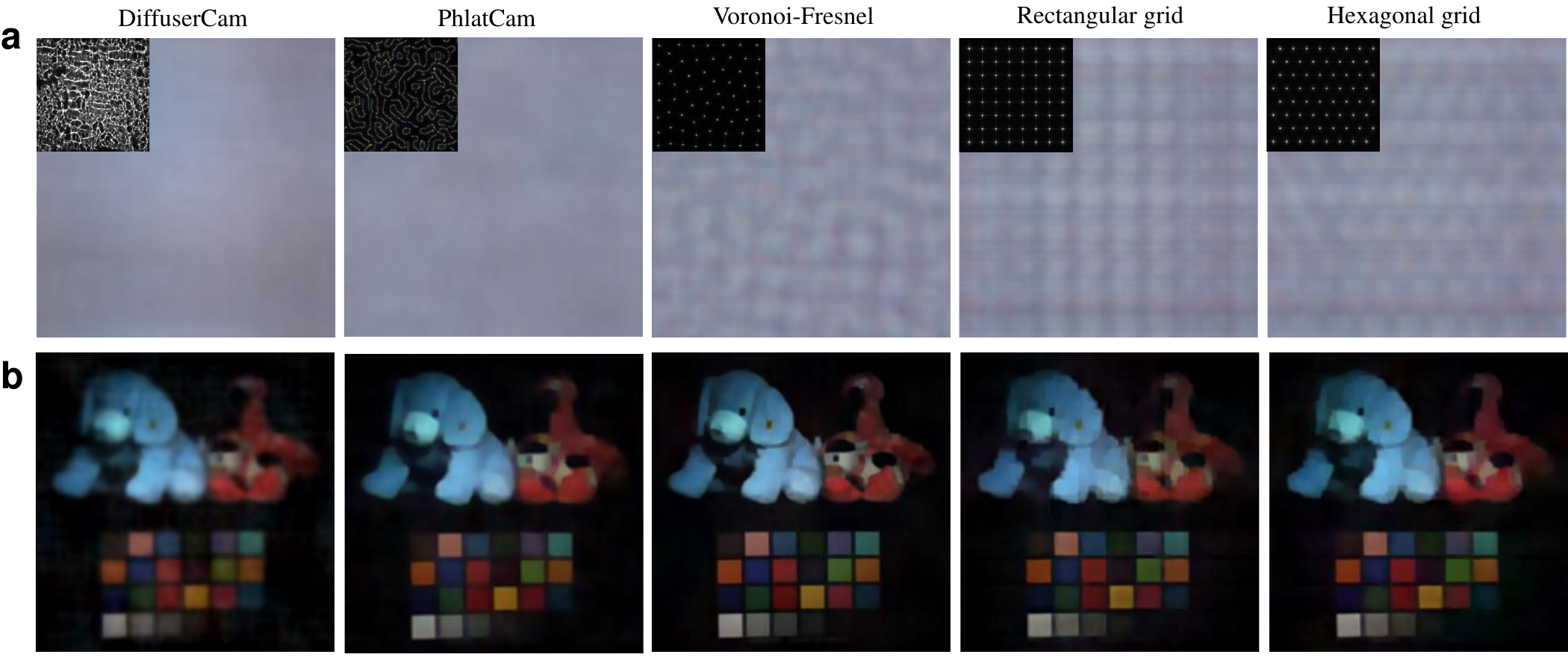}
    \caption{Visual Comparison of image reconstruction results. \change{(a) Raw sensor data for the (from left to right) DiffuserCam, PhlatCam, Voronoi-Fresnel, rectangular Fresnel, and hexagonal Fresnel. The corresponding PSFs are shown in the top-left corner. (b) Corresponding image reconstruction results.}}
    \label{fig:sim_results}
\end{figure*}

\begin{figure}[h]
    \centering
    \includegraphics[width=0.7\textwidth]{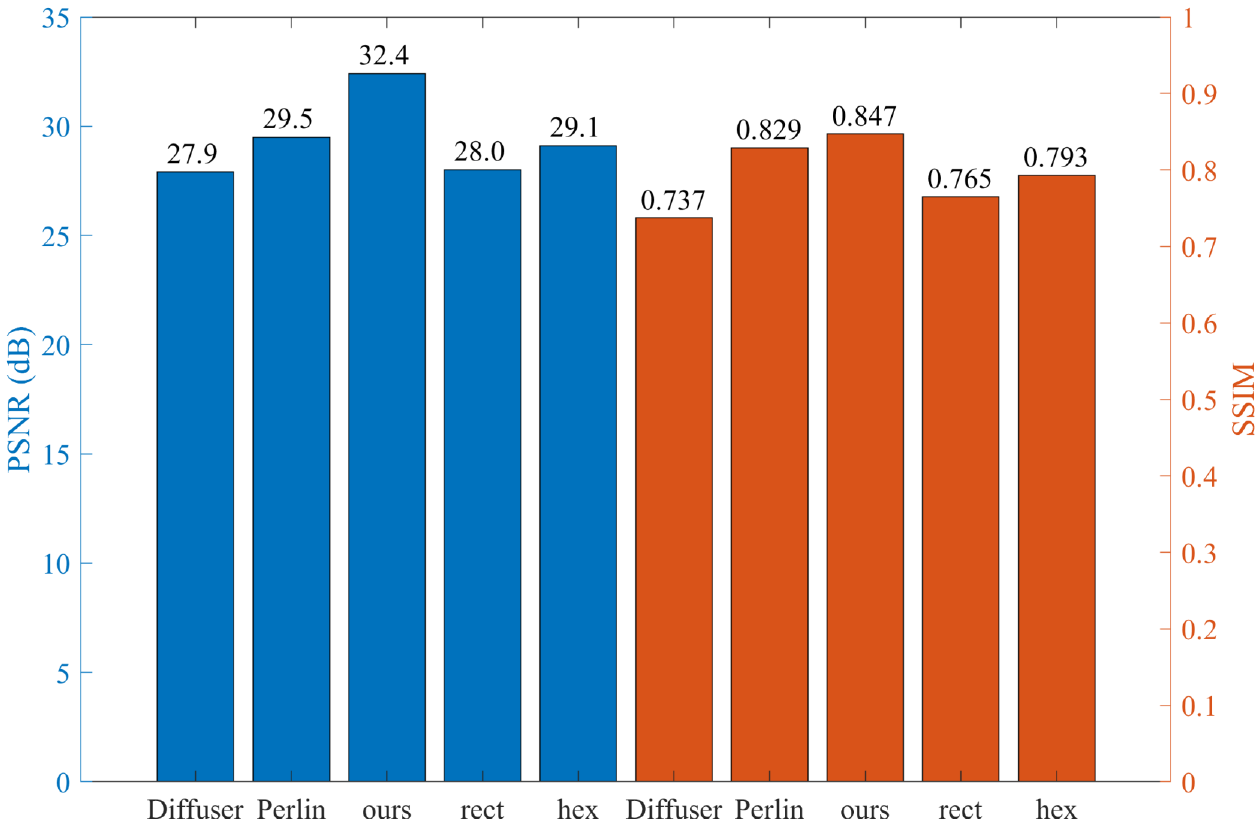}
    \caption{\change{Image reconstruction quality comparison in terms of PSNR and SSIM for the DiffuserCam, PhlatCam, optimized Voronoi-Fresnel (ours), rectangular Fresnel, and hexagonal Fresnel designs.}}
    \label{fig:sim_quality}
\end{figure}

An example raw data with their respective PSFs are shown in Fig.~\ref{fig:sim_results}a. The raw data from our lensless camera is more structured rather than flattened in the other two cases. \change{For a complete comparison, here we also include the regular Fresnel array configurations in both rectangular grid and hexagonal grid, as shown in Fig.~\ref{fig:properties}.} The corresponding reconstructed images are shown in Fig.~\ref{fig:sim_results}b. \change{We evaluate the performance by Peak Signal to Noise Ratio (PSNR) and Structural Similarity Index (SSIM) in Fig.~\ref{fig:sim_quality}.} The results show that, our Voronoi-Fresnel design outperforms existing methods in both PSNR and SSIM, owing to the MTFv guided optimization. \change{The optimized Voronoi-Fresnel configuration also performs better than the two reference regular Fresnel configurations.}

\subsection{Experimental results}

\subsubsection{Prototype design}
\begin{figure*}[h!]
    \centering
    \includegraphics[width=\textwidth]{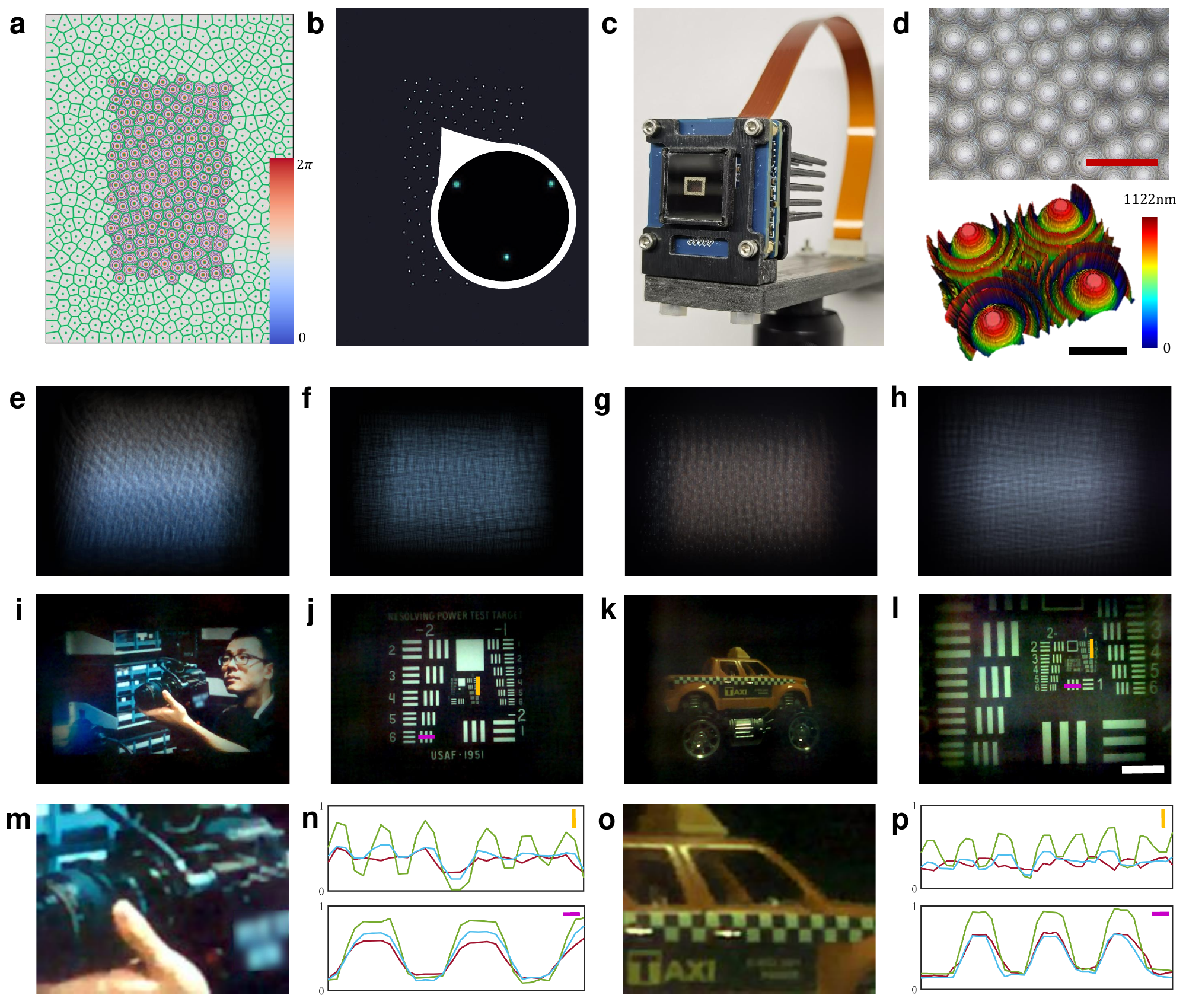}
    \caption{Experimental results with the prototype Voronoi-Fresnel lensless camera. (a) Optimized phase profile with a FOV of $\pm$20$^{\circ}$. Cell boundaries are marked by the overlaid green polygons. Marginal Voronoi-Fresnel cells are excluded to maintain the shift-invariance of the PSF. (b) Calibrated color PSF (brightness enchanced) with the magnified area showing three adjacent spots. (c) The prototype consists of a Voronoi-Fresnel phase element mounted with 3D-printed parts onto a board-level sensor. (d) Top is a 5$\times$ microscopic image of the fabricated 16-level sample on a Nikon Eclipse L200N microscope. Scale bar is 300 $\mu m$. Bottom is a 20$\times$ microstructure measurement on a Zygo NewView 7300 profilometer. Scale bar is 100 $\mu m$. (e) and (f) are the raw data for the scenes displayed on a computer monitor, and (i), (j) are the reconstructed images respectively. (m) is the zoom-in area in (i), and (n) represents the intensity profile along the colored  lines in USAF resolution chart in reconstruction (j). (g) and (h) represent the raw data the real objects taken under ambient illumination. The reconstructed images are (k) and (l) respectively. (o) is the zoom-in area in (k), and (p) shows the corresponding cross line in the real resolution chart in (l). \change{The photograph used in (i) is by courtesy of Jinhui Xiong with permission. The camera logo in (i) and (m) is intentionally blurred to avoid copyright infringement.}} 
    \label{fig:prototype}
\end{figure*}

A prototype Voronoi-Fresnel lensless camera is built with a board-level camera FLIR BFS-GE-16S2C-BD2 and a custom fabricated phase element. The image sensor (Sony IMX273) has 1440~$\times$~1080 pixels, with a pixel pitch of 3.45~$\mu$m. The optical distance in-between is 3~mm, where the cover glass on the sensor has a thickness of 0.7~mm. The sensor's angular response covers approximately $\pm \mathrm{20}^{\circ} \times \pm \mathrm{15}^{\circ}$ FOV, so the shift-invariance of the PSF should be maintained within this angular range. Marginal cells beyond the FOV are excluded for this purpose. The optimized phase profile is shown in Fig.~\ref{fig:prototype}a. The excluded cells are empty of the Fresnel phase. A calibration PSF is shown in Fig.~\ref{fig:prototype}b with a zoom-in area highlighting three adjacent spots. The optical element is fabricated on a 0.5-mm-thick fused silica substrate by a combination of photolithography and reactive-ion etching techniques. At a nominal wavelength of 550~nm, the $2\pi$ phase modulation corresponds to a total depth of 1200~nm, which is discretized into 16-levels for fabrication. The lateral fabrication resolution is 1.15~$\mu$m, making each sensor pixel upsampled by 3~$\times$~3 of the optical pixel. The optical element is assembled onto the sensor by 3D-printed mechanical mounts (Fig.~\ref{fig:prototype}c). An optical microscope image (Nikon Eclipse L200N, 5$\times$) and a 3D measurement (Zygo NewView 7300, 20$\times$) of the fabricated sample are shown in Fig.~\ref{fig:prototype}d. Details of the prototype design and fabrication can be found in the \tocheck{Supplemental Document 1}.

\subsubsection{Prototype results}
In the following, we present the raw data in full sensor resolution, and the reconstructed images are cropped to 640~$\times$~480 pixels to cover the $\pm \mathrm{20}^{\circ} \times \pm \mathrm{15}^{\circ}$ FOV. The object distance is about 30~cm away from the lensless camera. The exposure time is set to make sure all pixels not saturated ($\sim$ 240/255) in 8-bit mode. Gamma correction is disabled in the capture, and raw data are used in the reconstruction.

We evaluate the image performance in two illumination scenarios. The first is self-illuminating images displayed on a computer monitor. Self-illuminating objects emit light in a confined angular range, so little stray light or cross-talk is introduced in the captured data. Fluorescence imaging would be a good application for this mode. Figure~\ref{fig:prototype}e and \ref{fig:prototype}f show the captured raw data. The reconstructed image in Fig.~\ref{fig:prototype}i reveals fine details in the camera and human face. A zoom-in area of the camera object is shown in Fig.~\ref{fig:prototype}m. We evaluate the spatial resolution with a USAF resolution target in Fig.~\ref{fig:prototype}j, with two cross lines (yellow and magenta) plotted in Fig.~\ref{fig:prototype}n for the RGB color channels. Line pairs in the central area are clearly visible. 

The second is real objects with ambient illumination (Fig.~\ref{fig:prototype}g and \ref{fig:prototype}h). This is a more realistic scenario for photography outside of a lab environment or in applications like endoscopy. Ambient illumination poses a severe challenge for existing flat or lenseless imaging systems, since ``stray'' light can enter the camera at angles outside the nominal field of view, which is then not modeled by the reconstruction algorithm. We show the reconstructed car toy image example in Fig.~\ref{fig:prototype}k and a printed USAF resolution chart in Fig.~\ref{fig:prototype}l. Since the ambient illumination is not uniform across the scene, the intensity fall-off from on-axis to off-axis, which is vignetting, is more obvious than in the self-illuminating case. The recovered image is still able to reveal sufficient details despite of the complicated environmental light condition. Similarly we evaluate the spatial resolution with the printed USAF resolution target. The cross lines in Fig.~\ref{fig:prototype}p indicate that the line pairs can be discriminated very well in green and blue channels, while it becomes less reliable in the red channel in the high frequency line pairs. The differences in color channels also indicate residual chromatic aberrations exist in the reconstructed images. We present a more thorough discussion of the prototype, including characteristic measurements, color reproduction, resolution analysis, and additional results in \tocheck{Supplemental Document 1}.

\section{Discussion}
\subsection{Advantages over existing designs}
The proposed Voronoi-Fresnel lensless camera differentiates itself from existing lensless technologies that focus on either biomimetic optics or PSF engineering. Our design not only makes use of the compound-eye structure optically, but also facilitates the subsequent algorithm computationally. The compositing cells in our design do not work individually as in compound eyes, but form an optimized PSF collectively. Optical cross-talk between adjacent cells is not a concern, so no blocking layer is necessary. The reconstruction algorithm is not to stitch sub-images, but to solve an inverse problem from the physical model. Most prominently, unlike the regular tessellations, our method features an optimized irregular structure. The number of cells is determined by the imaging performance, not by the final image pixel counts.

Our Voronoi-Fresnel phase is mostly inspired by phase-only designs for PSF engineering, but pushes the optimality of the PSF structure further. The resulting PSF not only possesses the properties assumed in existing works, but also exhibits some unique features, such as more compact directional filtering, non-trivial random yet uniform distribution, and an optimal number of diffraction limited spots. As we have demonstrated, the distribution of the focusing units matter significantly. A search on the number of units is also necessary to find the optimal solution. Another related optimization metric is the auto-correlation of the PSF as used in Miniscope3D~\cite{yanny2020miniscope3d}. Since auto-correlation is not a single number, it cannot be used directly as an optimization metric. A diffraction limited MTF is necessary as the reference instead. We show a simple example in \tocheck{Supplemental Document 1} that, without a reference, this metric could lead to ambiguity. In contrast, our MTFv concept distills the spatial and spectral information in PSF into a single figure-of-merit that can be used for numerical optimization. MTFv requires no reference value, and evaluates the amount of useful information by itself. \change{Different from the custom MLAs in computational miniature mesoscope~\cite{xue2020single,xue2022deep} and learned 3D lensless camera~\cite{tian2022learned}, the number and geometries of our Voronoi-Fresnel cells are determined automatically by the algorithm.}

Additional benefits of the proposed design lie on various aspects. Since our design is a tessellation of the base Fresnel phase, it facilitates large-area design in high spatial resolution, which alleviates the sampling load for pixel-wise phase element optimization on mega-pixel sensors. The resulting smooth microstructures ease fabrication requirements than otherwise high-frequency random features in prior designs. The Voronoi-Fresnel design finds an in-between strategy that takes advantages of both compound-eye and PSF engineering methods.

\subsection{Optical characteristics}
\begin{figure}[h!]
    \centering
    \includegraphics[width=\columnwidth]{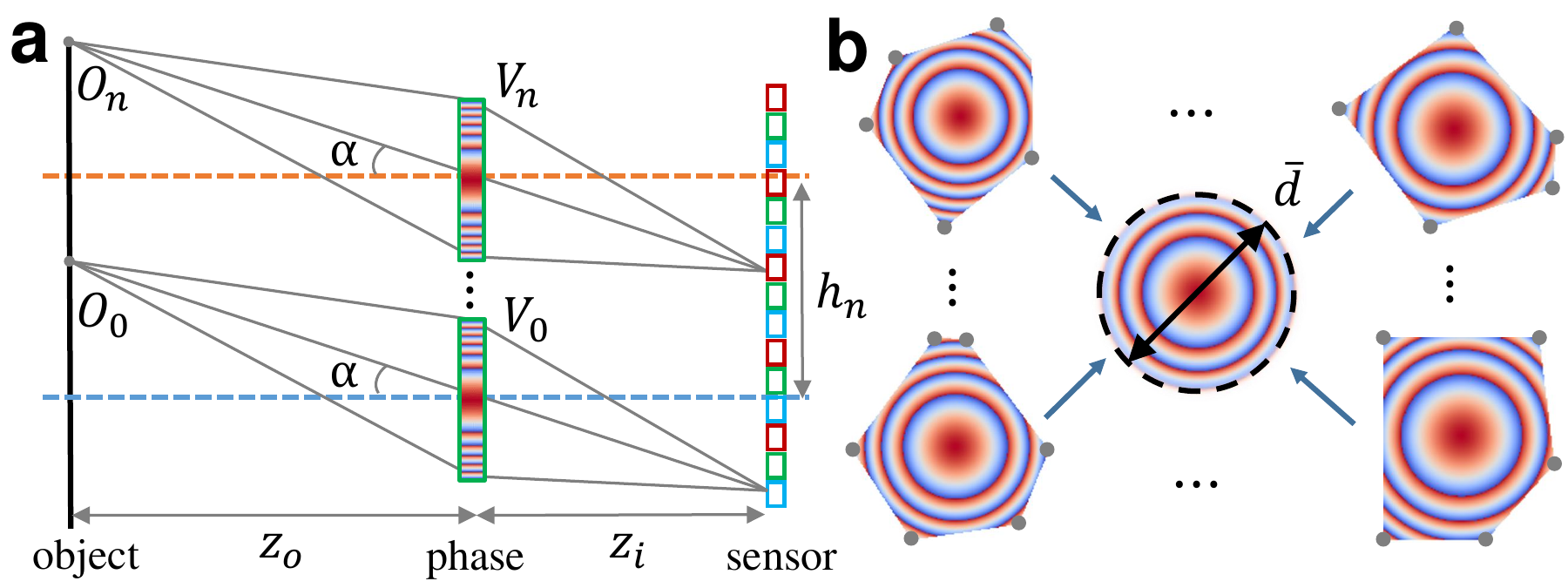}
    \caption{Field-of-view and limiting resolution. (a) The FOV is determined by the sensor cut-off angle $\alpha$, marginal Voronoi-Fresnel cell ($V_n$) displacement $h_n$, the object distance $z_o$, and phase-to-sensor distance $z_i$. (b) The statistical diameter of the base Fresnel phase $\bar{d}$ is the RMS distance from the vertices to their respective center locations.}
    \label{fig:discussion}
\end{figure}

Compared to its lens counterparts, the performance of the Voronoi-Fresnel lensless imaging can also be characterized and analyzed by effective focal length, spatial resolution, FOV and so on. Since the Voronoi-Fresnel phase is a collection of the same base Fresnel phase, the effective focal length is also the focal length of the base Fresnel phase. The FOV is determined by three factors, as shown in Fig.~\ref{fig:discussion}a. First, the image sensor usually has a cut-off angle $\alpha$, beyond which light is not sensible. Each Voronoi-Fresnel cell is limited by this cut-off angle. The outmost object that the central cell $V_{0}$ can see is $O_{0}$ in the object space. Second, the marginal cell has a lateral center displacement $h_{n}$ from the optical axis. This corresponds to an angular displacement of $ \arctan{\left( h_{n} / z_{o} \right) } $. The equivalent half FOV is then
\begin{equation}
\theta = \alpha + \arctan{\left( \frac{h_{n}}{z_{o}} \right)} = \alpha + \arctan{\left( \frac{m h_{n}}{z_{i}} \right)},
\end{equation}
where $m = z_i / z_o$ is the magnification calculated from the object distance $z_o$, and phase-to-sensor distance $z_i$. In our experiments we did not remove the sensor cover glass, which impose a minimum constraint on this distance.

The resolution of lensless cameras is usually object-dependent. Since the base Fresnel phase is a first-order approximation of an ideal lens, the individual cell is closely diffraction-limited. There exists a theoretical limiting resolution. However, variations in the aperture shapes do exist between different Voronoi-Fresnel cells. We define an effective diameter $\bar{d}$ for the base Fresnel phase by statistically calculating the Root-Mean-Square (RMS) distance of all the vertices to their respective center locations. Assuming the $i$-th cell has $M_{i}$ vertices, and there are $N$ cells in total, the RMS diameter is
\begin{equation}
\bar{d} = 2 \sqrt{\frac{\sum_{i=1}^{N} \sum_{j=1}^{M_i} \left(x_{i}^{j} - x_{i}\right)^2 + \left(y_{i}^{j} - y_{i}\right)^2 }{\sum_{i=1}^{N} M_i}},
\end{equation}
where $\left(x_i, y_i\right)$ are the $i$-th center coordinates, and $(x_{i}^{j}, y_{i}^{j})$ are the $j$-th vertex in the $i$-th cell. The equivalent limiting resolution can be evaluated according to the conventional Rayleigh criterion, i.e., the radius of the distinguishable spot is $1.22 \lambda z_{i} / \bar{d}$. For our prototype, the effective diameter $\bar{d}$ is 233.5~$\mu$m, so the diffraction limited spot diameter is 15.7~$\mu$m. We evaluate the prototype resolution by imaging a cross-hair target, and measure the cross-section diameters. The measured resolution is 18.4~$\mu$m and 20.7~$\mu$m in horizontal and vertical directions respectively (see \tocheck{Supplemental Document 1}). In practice, the final image quality is degraded by the complexity of the scene, the noise, and the algorithm efficacy. Owing to the blue-noise sampling, the optimized MTF shows minimal low-frequency components, while keeping mid-frequency and high-frequency well. This indicates the contrast in smooth objects is sacrificed to allow more mid-frequency and high-frequency information to be recorded.

\subsection{Depth-dependent PSFs}
\begin{figure}[h!]
    \centering
    \includegraphics[width=\textwidth]{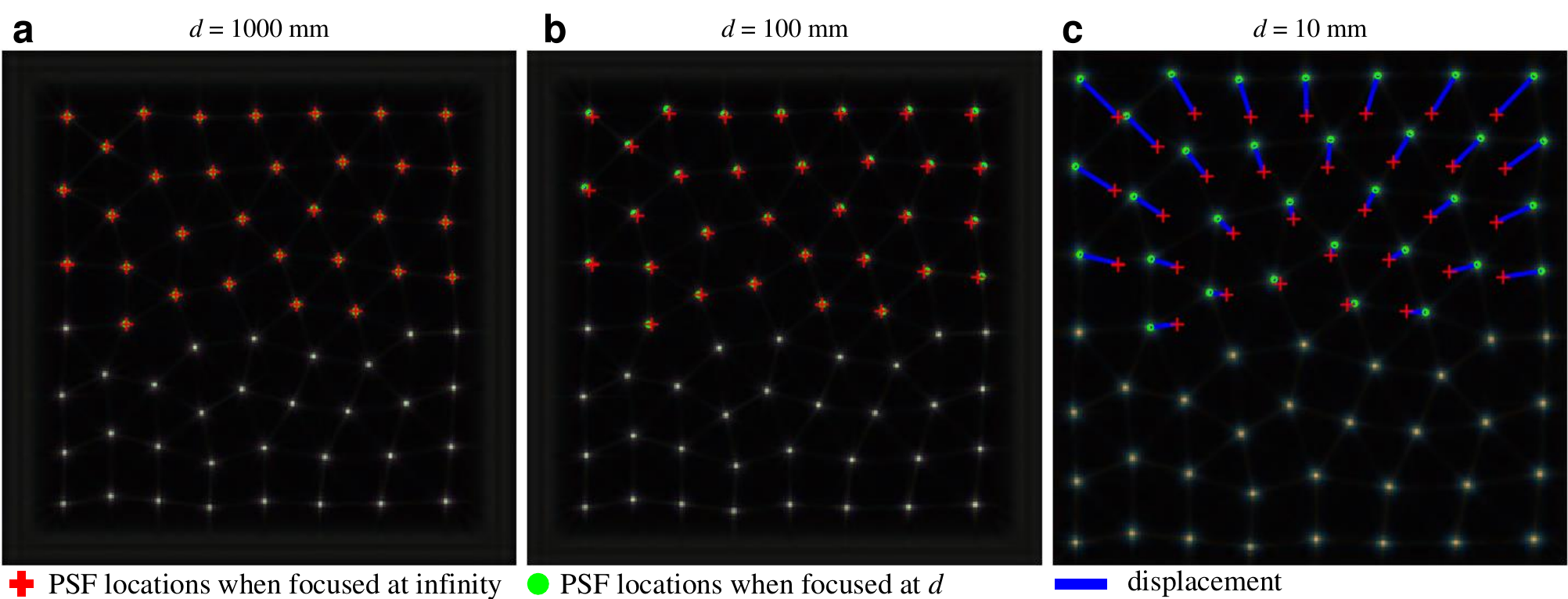}
    \caption{PSF expansion with object distances. We place the object at (a) 1000~mm, (b) 100~mm, and (c) 10~mm respectively to observe the displacements of the individual spot locations. Red cross symbols indicate the reference locations when focused at infinity, while the green dot symbols are for near object distances $d$. The blue lines denote the corresponding displacements. Only the top half of the spots are labeled here for brevity.}
    \label{fig:psf_depth}
\end{figure}

Similar as the PSFs in DiffuserCam~\cite{antipa2018diffusercam} and PhlatCam~\cite{boominathan2020phlatcam}, our PSF expands laterally when the depth is closer to the sensor, while keeping nearly constant when the object is far away. \change{To demonstrate the PSF expansion, we simulate the depth varying PSFs for the optimized example in Fig~\ref{fig:properties}c. The results are shown in Fig.~\ref{fig:psf_depth}. We label the centroids of the spots in the PSF at infinity in red as a reference. The centroids of the PSFs at the evaluated depths are shown in green, and the corresponding displacements are denoted with blue lines.} It can be seen that the spot locations displace much more for the small object distance of $d = 10$~mm than larger distances of 100~mm and 1000~mm, with respect to the infinity PSFs. \change{The lateral displacements of the spots make the PSFs at different depths distinct and less correlated with each other.}

\begin{figure}[h!]
    \centering
    \includegraphics[width=\textwidth]{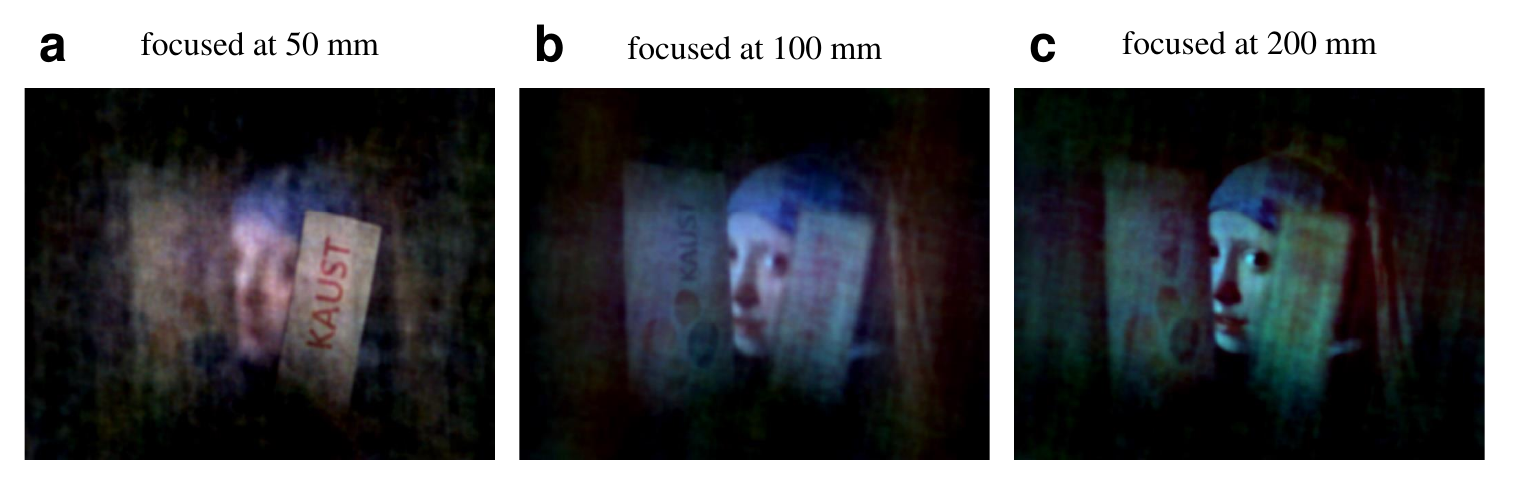}
    \caption{\change{Prototype imaging results for 3D scenes. (a) Reconstructed image focusing at near object (50~mm). (b) Reconstructed image focusing at middle object (100~mm). (c) Reconstructed image focusing at far object (200~mm). The logo and letters are used with permission by KAUST.}}
    \label{fig:prototype_depth}
\end{figure}

\change{Since the PSFs are depth-dependent, our Voronoi-Fresnel lensless camera is also capable of imaging 3D scenes. To demonstrate this, we show in Fig.~\ref{fig:prototype_depth} an example of refocusing at various depths with the prototype. We place three objects in front of the lensless camera at 50~mm, 100~mm and 200~mm, respectively. With one single shot, we can reconstruct the corresponding images that are focused at different depths. For example, in Fig.~\ref{fig:prototype_depth}a it is focused at the letters ``KAUST'' at 50~mm, and the rest of the scene are severely blurred. In Fig.~\ref{fig:prototype_depth}b the logo in the middle (100~mm) is focused, and the front letters are completely unrecognizable, while the drawing in the back is slightly blurred. Similarly when the drawing in Fig.~\ref{fig:prototype_depth}c is sharp, the other two objects are out-of-focus.}

\change{We also note that, in 2D photography applications the PSFs are often nearly constant. To account for depths, not only geometric displacement in the PSFs but also their spectral variations could be employed to encode depths. The above example only shows the geometric effect. It is also possible to optimize the base phase in each cell to make the PSFs more suitable for 3D applications. Furthermore, in the reconstruction, this requires depth-aware optimization of the phase together with the reconstruction scheme. The end-to-end optimization in a deep learning framework would be explored in future work.}

\subsection{Limitations and future work}
The proposed Voronoi-Fresnel lensless camera also exhibits a few limitations. First, the two-step optimization algorithm is time-consuming due to the parameter sweep step. Although the scalability property helps estimating the correct range, it is still expected to develop more efficient algorithms to accelerate the process for large-area designs. Second, the FOV is currently limited by the cut-off angle of the sensor, which is mainly due to the microlens array embedded in the image sensor. The intensity fall-off may be a concern for wide-angle applications. It may be possible to design telecentric Voronoi-Fresnel phases to mitigate the situation, although fabrication complexity would be increased. Another limitation for the current prototype is the stray light. A common practice in conventional lenses to control stray light is to use baffles. However, it is difficult to implement such an idea without sacrificing the compactness. In the apposition type compound eyes, evolution has developed a screening pigment around the ommatidium to block stray light and cross-talk from entering the photoabsorbing rhabdom. Similar structures can be fabricated by etching deep trenches around the boundaries of Voronoi-Fresnel cells, and fill in black materials. Finally, an end-to-end framework that incorporates the optical element to be optimized together with the reconstruction network is still lacking, and is worth exploring in future work.

\section{Conclusion}
We have demonstrated a compound eye inspired lensless camera with optimal information preservation in optics. Following the proposed Fourier domain metric MTFv, we are able to tailor a spatially-coded Voronoi-Fresnel phase for better computational image reconstruction. Experimental results show the superior image quality of the prototype lensless camera in various illumination conditions. The advantages of the proposed Voronoi-Fresnel lensless camera offer a simple yet cost-effective imaging solution to significantly reducing the volume of imaging devices. The possibility of mass production makes it a promising candidate in applications such as fluorescence imaging, endoscopy, and internet-of-things.

\begin{backmatter}
\bmsection{Funding}
King Abdullah University of Science and Technology (Individual Baseline Funding); National Natural Science Foundation of China (62172415); National Key Research and Development Program of China (2019YFB2204104).

\bmsection{Acknowledgments}
This work was partly done in the Nanofabrication CoreLabs (NCL) at KAUST. 

\bmsection{Disclosures}
The authors declare no conflicts of interest.

\bmsection{Data availability} Data underlying the results presented in this paper are not publicly available at this time but may be obtained from the authors upon reasonable request.

\bmsection{Supplemental document}
See Supplemental Document 1 for supporting content. 

\end{backmatter}


\bibliography{ref}






\end{document}


\maketitle

\section{Point Spread Function}
Here we derive the mathematical expressions for the point spread functions (PSFs) in the main texts. First, we inspect a single Voronoi-Fresnel cell $V_i$. The complex optical field after propagating from the phase element to the image sensor plane is
\begin{equation}
\begin{aligned}
&P_i \left( x, y, \lambda \right) =\Fr{ A_{i} \left(\xi - \xi_{i}, \eta - \eta_{i} \right) \exp \left(- j \frac{2\pi}{\lambda} \frac{\left(\xi - \xi_{i}\right)^2 + \left(\eta - \eta_{i}\right)^2}{2z} \right) }\\
&= \frac{1}{j \lambda z} \iint A_{i} \left(\xi', \eta' \right) \exp \left(- j \frac{\pi}{\lambda z} \left(\xi'^2 + \eta'^2 \right) \right) \exp \left( j \frac{\pi}{\lambda z} \left( \left(x-\xi_i-\xi'\right)^2 + \left(y-\eta_i-\eta'\right)^2 \right) \right) \mathrm{d} \xi' \mathrm{d} \eta'\\
&=\frac{\exp \left( \frac{\pi}{\lambda z} \left( \left(x - \xi_i\right)^2 + \left( y - \eta_i\right)^2 \right) \right)}{j \lambda z} \iint A_{i} \left(\xi', \eta' \right) \exp \left( - j \frac{2 \pi}{\lambda z} \left( \left(x - \xi_i\right) \xi' + \left(y - \eta_i\right) \eta' \right) \right) \mathrm{d} \xi' \mathrm{d} \eta'\\
&= \frac{\exp \left( \frac{\pi}{\lambda z} \left( \left(x - \xi_i\right)^2 + \left( y - \eta_i\right)^2 \right) \right)}{j \lambda z} \mathcal{F} \left\{ A_{i} \left(\xi', \eta' \right) \right\} \left|_{\left[ \frac{x-\xi_i}{\lambda z}, \frac{y-\eta_i}{\lambda z} \right]} \right.\\
&=\frac{C \left(\xi_{i}, \eta_{i} \right)}{j \lambda z} \ft{ A_{i} \left(\xi', \eta' \right) } \left|_{\left[ \frac{x-\xi_i}{\lambda z}, \frac{y-\eta_i}{\lambda z} \right]} \right.,
\end{aligned}
\label{eq:supp_diffraction}
\end{equation}
where we have replaced the variables by $\xi' = \xi - \xi_i$ and $\eta' = \eta - \eta_i$ in the second line. The integral in the third line is exactly the Fourier transform of the aperture function $A_i$ evaluated at spatial frequencies $\left[ \left(x-\xi_i\right) / \lambda z, \left(y-\eta_i\right) / \lambda z \right]$. The corresponding PSF is then
\begin{equation}
\begin{aligned}
\psf_i \left(x, y, \lambda\right) \left(x, y, \lambda\right) &= \left| P_i \left(x, y, \lambda\right) \right|^2\\
&\propto \left| \mathcal{F} \left\{ A_{i} \left(\xi', \eta' \right) \right\} \left|_{\left[ \frac{x-\xi_i}{\lambda z}, \frac{y-\eta_i}{\lambda z} \right]} \right. \right|^2\\
&= \psf_i^0 \left(x - \xi_i, y - \eta_i, \lambda \right),
\end{aligned}
\end{equation}
where we denote $\psf_i^0 \left(x, y, \lambda \right)$ as the centered PSF as if the aperture were located at the origin,
\begin{equation}
\begin{aligned}
\psf_i^0 \left(x, y, \lambda \right) &\propto \left| \mathcal{F} \left\{ A_{i} \left(\xi', \eta' \right) \right\} \right|^2 \\
&= \left| \iint A_{i} \left(\xi', \eta' \right) \exp \left( - j \frac{2 \pi}{\lambda z} \left( x \xi' + y \eta' \right) \right) \mathrm{d} \xi' \mathrm{d} \eta' \right|^2.
\end{aligned}
\end{equation}
This implies that the PSF for the $i$-th Voronoi-Fresnel cell is a shifted version of the centered PSF. It is worth noting that the shape and distribution of the centered PSF depend on the geometry of the aperture functions. These apertures are finite-edge polygons, so the centered PSFs are actually compact yet highly directional filters. It is important to diversify such directional filtering of the PSFs to achieve optimal performance.

The effective PSF of the Voronoi-Fresnel lensless camera is obtained in the same way by taking the whole phase into account,
\begin{equation}
\begin{aligned}
\psf \left(x, y, \lambda\right) &= \left| \Fr{\Phi \left(\xi, \eta, \lambda \right)} \right|^2\\
&= \left| \Fr{\sum_{i=1}^{K} A_{i} \left(\xi - \xi_{i}, \eta - \eta_{i} \right) \cdot \exp{\left( -j \frac{2\pi}{\lambda} \cdot \frac{\left(\xi - \xi_{i}\right)^2 + \left(\eta - \eta_{i}\right)^2}{2z}\right)}} \right|^2\\
&= \left| \sum_{i=1}^{K} P_i \left( x, y, \lambda \right) \right|^2\\
&= \left( \sum_{i=1}^{K} P^*_i \left( x, y, \lambda \right) \right) \left( \sum_{i=1}^{K} P_i \left( x, y, \lambda \right) \right)\\
&= \sum_{i=1}^{K} \left| P_i \left( x, y, \lambda \right) \right|^2 + \sum_{
j = 1
\atop
i \ne j
}^{K} \sum_{i=1}^{K} P^*_i \left( x, y, \lambda \right) P_j \left( x, y, \lambda \right).
\end{aligned}
\label{eq:supp_cross}
\end{equation}
Although the Voronoi cells share no intersections, $V_{i} \cap V_{j} = \emptyset, \, \forall i \neq j$, and the individual aperture functions have no overlapped areas, the diffraction patterns $P_{i}\left(x, y, \lambda\right)$ and $P_{j}\left(x, y, \lambda\right)$ in general would interfere with each other, so the cross terms in ~\eqref{eq:supp_cross} are not necessarily zero. 

\begin{figure}[ht]
\centering
\includegraphics[width=\textwidth]{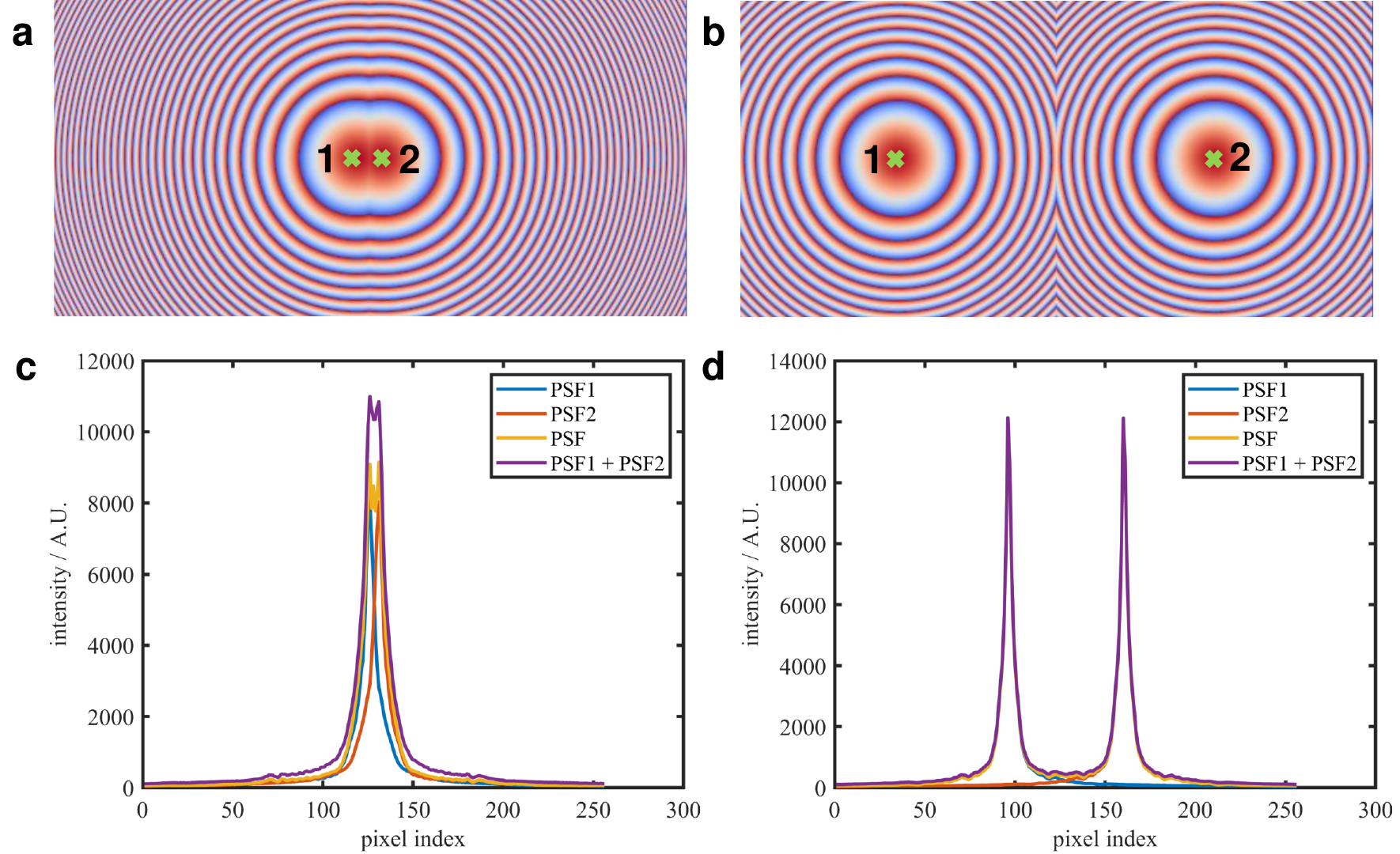}
\caption{Cross terms in two adjacent PSFs. (a) Two cells with very close centers near the boundary. (b) Two cells with centers at the centroids. (c) Cross-section of the individual PSFs, total PSFs and the sum of individual PSFs in (a). (d) Cross-section of the individual PSFs, total PSFs and the sum of individual PSFs in (b). Maximum error is less than 1\% when the the centers are at the centroids of the two cells.}
\label{fig:PSF_cross}
\end{figure}

We investigate the cross terms in two situations. One is a random distribution of adjacent Fresnel centers, and the other is where the centers are at the centroids. A simple example is a rectangular space tessellated with two adjacent cells, as shown in Fig.~\ref{fig:PSF_cross}. The optical parameters are the same as in \tocheck{Fig.~3} in the main paper. When the centers of two adjacent cells are not located in the centroids, and very close to each other near the boundary, the sum of individual PSFs are different from the real PSF predicted by the Fresnel propagation. However, when the centers are at the centroids of the cells, the sum of individual PSFs is approximately the same as the accurate PSF predicted by the Fresnel diffraction. The maximum error is less than 1\%.

Hence, for the Centroidal Voronoi case, the entire PSF can be approximated by omitting the cross terms, just for the purpose of analysis,
\begin{equation}
\psf \left(x, y, \lambda\right) \approx \sum_{i=1}^{K} \left| P_i \left( x, y, \lambda \right) \right|^2 = \sum_{i=1}^{K} \psf_i^0 \left(x - \xi_i, y - \eta_i, \lambda\right),
\end{equation}
Note that in the simulation and optimization, we do not simulate individual PSFs and superposition them, but use the entire constructed phase function to obtain the panchromatic PSF.

The above PSF is for monochromatic light. To get the panchromatic PSF, we simply integrate all the spectral PSFs over the interested spectral range, 
\begin{equation}
\psf \left(x, y\right) = \int_{\lambda_1}^{\lambda_2} \psf \left(x, y, \lambda\right) \mathrm{d} \lambda.
\end{equation}

\section{Modulation Transfer Function}
Here we provide a detailed derivation and analysis for the Modulation Transfer Function (MTF). MTF is defined as the magnitude of the Optical Transfer Function (OTF), which is the Fourier transform of the PSF for incoherent imaging systems,
\begin{equation}
\mtf = \left| \otf \right| = \left| \ft{\psf}  \right|,
\end{equation}
where $0 \leq \mtf \leq 1$. We can obtain the MTF by taking the Fourier transform of the above PSF,
\begin{equation}
\begin{aligned}
\mtf \left(f_X, f_Y\right) &= \left| \ft{\psf \left(x, y \right)} \right|\\
&\approx \left| \sum_{i=1}^{K} \ft{\psf_{i}^0 \left(x - \xi_i, y - \eta_i\right)} \right|\\
&= \left| \sum_{i=1}^{K} \ft{\psf_{i}^0 \left(x, y \right)} \exp \left(- j f_X \xi_i, - j f_Y \eta_i \right) \right|\\
&= \left| \sum_{i=1}^{K} \otf_{i}^0 \left(f_X, f_Y \right) \exp \left(- j f_X \xi_i, - j f_Y \eta_i \right) \right|,
\end{aligned}
\end{equation}
where $f_X$ and $f_Y$ are the Fourier domain frequencies, and we have applied the translation property of Fourier transform. The individual OTFs in the complex form are 
\begin{equation}
\otf_{i}^0 \left(f_X, f_Y\right) = \mathcal{M}_i \left(f_X, f_Y\right) \exp \left( -j \mathcal{P}_i \left(f_X, f_Y\right)\right).
\end{equation}
The remaining phase delay terms are simplified as
\begin{equation}
\exp \left( - j \Psi_i \left(f_X, f_Y\right) \right) = \exp \left(- j f_X \xi_i, - j f_Y \eta_i \right),
\end{equation} 
so the MTF can now be rewritten as
\begin{equation}
\mtf  \left(f_X, f_Y \right) = \left| \sum_{i=1}^K \mathcal{M}_i \left(f_X, f_Y\right) \exp \left(-j \mathcal{P}_i \left(f_X, f_Y\right) \right) \exp \left(-j \Psi_i \left(f_X, f_Y\right) \right) \right|.
\label{eq:supp_mtf}
\end{equation}
This equation reveals three factors that affect the MTF, the diffraction by each similar apertures that determines the $ \mathcal{M} \left(f_X, f_Y\right) $ and $\mathcal{P}_i \left(f_X, f_Y\right)$ terms; the additional phase delay terms $ \Psi_i \left(f_X, f_Y\right)$ that is introduced by the amount of spatial shifts from the centered PSFs; and the total number of Voronoi-Fresnel cells $K$.

In addition, we show how MTFv is related to the Strehl ratio. Strehl ratio is defined as the peak intensity ratio between the aberrated PSF and the diffraction limited PSF~\cite{brigantic1997image},
\begin{equation}
\textrm{SR} = \frac{I_{\mathrm{ab}}(0,0)}{I_\mathrm{dl}(0,0)},
\end{equation}
where $I_{\mathrm{ab}}(0,0)$ is the peak intensity of the aberrated PSF in the origin, and $I_{\mathrm{dl}}(0,0)$ is the corresponding intensity of the diffraction limited PSF. In practice, it is difficult to use Strehl ratio in this form as an optimization metric, as it is challenging to optimize the peak intensity of the PSF, and there may be multiple peaks in a composite system like ours. According to the definition of Fourier transform, we can rewrite the above equation in the Frequency domain,
\begin{equation}
\textrm{SR} = \frac{\iint \otf_{ab} \left(f_X, f_Y\right) df_X df_Y}{\iint \otf_{dl} \left(f_X, f_Y\right) df_X df_Y}.
\end{equation}
Now it becomes a quantity that can be calculated more easily with OTFs. We also note that, since the diffraction limited OTF is in the denominator, and is often a fixed value for a given system. We only need the term in the numerator. In addition, OTF consists of complex values. For stable numerical computation, we can replace OTF with MTF in the above equation. Since MTF is positive definite, the integral of MTF is always no less than the integral of OTF~\cite{roberts2002characterization},
\begin{equation}
\iint \otf_{ab} \left(f_X, f_Y\right) df_X df_Y \leq \iint \mtf_{ab} \left(f_X, f_Y\right) df_X df_Y.
\end{equation}
So finally we have
\begin{equation}
\mtfv = \iint \mtf_{ab} \left(f_X, f_Y\right) df_X df_Y.
\end{equation}
We note that, strictly speaking, although MTFv is related to the Strehl ratio, they are not identical. MTFv is a more generalized term that can be calculated easily from the MTF. It is also a tractable quantity that measures the information collected in an optical system. Therefore, we choose MTFv as a figure-of-merit for the optimization of our lensless imaging system.

\section{Centroidal Voronoi Tessellation}
Finding the optimal tessellation of the 2D design space is basically a sampling problem. Among various sampling methods, blue noise sampling~\cite{yan2015survey,de2012blue} offers minimal low-frequency components and no concentrated spikes of energy, which is the required properties for our application. An effective way to achieve blue noise sampling is by Centroidal Voronoi Tessellation (CVT)~\cite{du1999centroidal}. The CVT is a special Voronoi diagram where the sites coincide with the mass centroids of the corresponding Voronoi regions. The mass centroid of a Voronoi region $V_i$ is defined as
\begin{equation}
\Vect{c}_i = \frac{\int_{V_i} \Vect{p} \tau \left(\Vect{p}\right) \mathrm{d} \Vect{p}}{\int_{V_i} \tau \left(\Vect{p}\right) \mathrm{d} \Vect{p}},
\label{eq:centroid}
\end{equation}
where $\Vect{p}$ is a point in the Cartesian coordinates, and $\tau$ is a given density function. We can assume a constant density across the 2D plane for simplicity, i.e., $\tau \equiv 1$. CVT is a critical point of the energy function defined by
\begin{equation}
E_{\mathrm{CVT}} \left( P \right) = \sum_{i=1}^{K} \int_{V_i} \tau \left(\Vect{p}\right) \left\Vert \Vect{p} - \Vect{p}_i \right\Vert^2 \mathrm{d} \Vect{p}.
\end{equation}

There are various algorithms to optimize the above energy function and generate optimal CVT~\cite{du1999centroidal,yan2011computing}. A classic method is to use Lloyd iterations~\cite{lloyd1982least} to update the Voronoi sites by their centroids until convergence. In each iteration, the mass centroids are computed for the current Voronoi regions. Then these generated sites are replaced by the calculated centroids, and a new Voronoi tessellation is constructed. The process is repeated until a convergence criterion is met. 

\section{Scalability}
\begin{figure*}[h!]
    \centering
    \includegraphics[width=\textwidth]{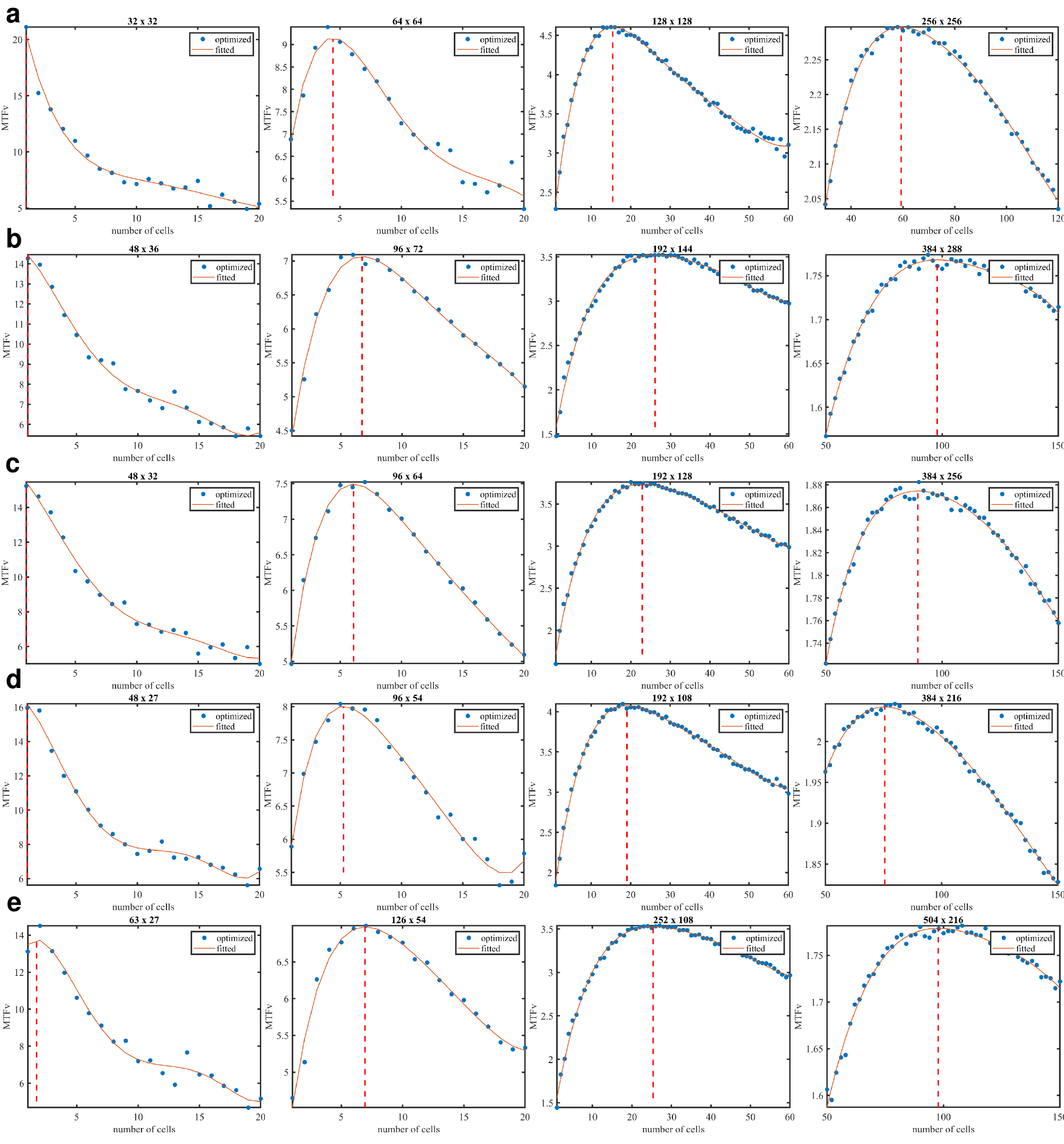}
    \caption{Linear scalability analysis for various aspect ratios. (a) Aspect ratio 1:1. (b) Aspect ratio 4:3. (c) Aspect ratio 3:2. (d) Aspect ratio 16:9. (e) Aspect ratio 21:9. For each aspect ratio, we quadruple the previous design area from left to right, as indicated in the figure titles (unit: $\mu m^2$).} 
    \label{fig:aspectratio}
\end{figure*}

We find that the optimal number of Voronoi-Fresnel cells is linearly proportional to the design area. Here we analyze and validate this assumption for different design areas with various aspect ratios. For all the experimental designs below, we assume the substrate is fused silica, and design the phase at 550~nm. The distance between the phase and sensor is 2~mm. Phase pixel size is 1~$\mu$m. In Fig.~\ref{fig:aspectratio}a-e we show the optimization curves for these experimental designs of aspect ratios of 1:1, 4:3, 3:2, 16:9, and 21:9, respectively. These aspect ratios allow us to account for common sensor shapes. For each aspect ratio, we start from a small area to optimize the Voronoi-Fresnel phase for various number of cells. Then we double the size in both dimensions (scaling to a quadruple area). We repeat this procedure 4 times for each aspect ratio. The best number of Voronoi-Fresnel cells in each design is obtained by fitting the data into a 5th order polynomial, and evaluate the cell number when the MTFv reaches the peak. These data are used in \tocheck{Fig.~5} in the main paper.

\section{Image Reconstruction}
A color image recorded on the image sensor is an integral of spectrally-blurred images weighted by the color response of the sensor. This process can be expressed as
\begin{equation}
g_c \left( x, y \right) = \int_{\lambda_1}^{\lambda_2} q_{c} \left( \lambda \right) \left( f \left(x, y, \lambda\right) * h \left(x, y, \lambda\right) \right) \mathrm{d} \lambda,
\label{eq:imageformation_spec}
\end{equation}
where $f \left(x, y, \lambda\right)$ is the latent spectral image at wavelength $\lambda$, $h \left(x, y, \lambda\right)$ is the spectral PSF, and $*$ denotes spatial convolution. $q_{c} \left( \lambda \right)$ is the color response function of the sensor, and $g_c \left( x, y \right)$ is the captured color image in color channel $c$ (for sensors with  Bayer filters, $c = r, g, b$). 

If the imaging system is perfect, we assume the spectral PSFs are all identically Dirac delta functions, i.e., $h \left(x, y, \lambda\right) = \delta \left(x, y, \lambda\right) = \delta \left(x, y\right)$. The ground-truth sharp image would be
\begin{equation}
f_c \left( x, y \right) = \int_{\lambda_1}^{\lambda_2} q_{c} \left( \lambda \right) f \left(x, y, \lambda\right) \mathrm{d} \lambda.
\end{equation}
On the other hand, if the image is a spectrally-uniform ideal point source, $f \left(x, y, \lambda\right) = \delta \left(x, y, \lambda\right) = \delta \left(x, y\right)$, the captured image would be a color PSF
\begin{equation}
h_c \left( x, y \right) =  \int_{\lambda_1}^{\lambda_2} q_{c} \left( \lambda \right) h \left(x, y, \lambda\right) \mathrm{d} \lambda.
\end{equation}

\begin{figure*}[h!]
    \centering
    \includegraphics[width=\textwidth]{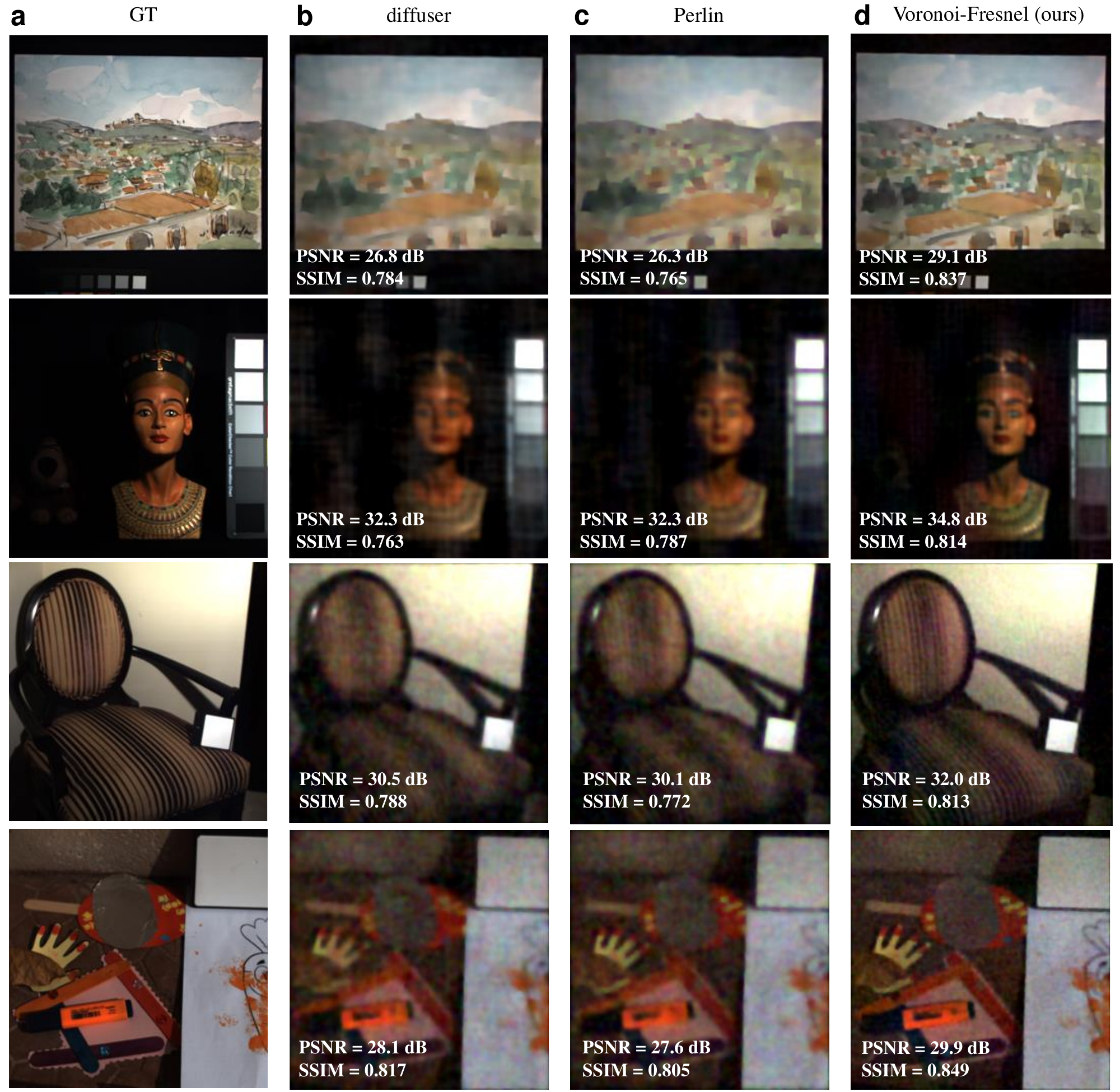}
    \caption{Additional comparison results. (a) Reference ground truth image. (b) Reconstructed image by the diffuser PSF. (c) Reconstructed image by the Perlin PSF. (d) Reconstructed image by the Voronoi-Fresnel PSF.}
    \label{fig:supp_simulation}
\end{figure*}

Note that if the imaging system is approximately achromatic, $h \left(x,y,\lambda\right) \approx h \left(x,y\right)$, \eqref{eq:imageformation_spec} can be simplified by
\begin{equation}
\begin{aligned}
g_c \left( x, y \right) &\approx \left( \int_{\lambda_1}^{\lambda_2} q_{c} \left( \lambda \right) f \left(x, y, \lambda\right) \mathrm{d} \lambda \right) * h \left(x, y\right)\\
&= f_c \left(x, y\right) * \int_{\lambda_1}^{\lambda_2} q_{c} \left(\lambda\right) h \left(x, y\right) \mathrm{d} \lambda\\
&= f_c \left(x, y\right) * h_c \left(x, y\right),
\end{aligned}
\end{equation}
where the sensor response is normalized such that $\int_{\lambda_1}^{\lambda_2} q_{c} \left( \lambda \right) \mathrm{d} \lambda = 1$. To solve a minimization problem in the ADMM framework, we introduce a slack variable $\Vect{z} = \Mat{D} \Vect{x}$, and apply the augmented Lagrangian multiplier,
\begin{equation}
\argmin_{\Vect{x}} \frac{1}{2} \left\Vert  \Mat{A} \Vect{x} - \Vect{y} \right\Vert^2_2 + \mu \left\Vert \Vect{z} \right\Vert_{1} + \frac{\rho}{2} \left\Vert \Mat{D} \Vect{x} - \Vect{z} \right\Vert^2_2,
\end{equation}
where $\rho$ is the weight for the slack variable. The ADMM iterations consist of three steps,
\begin{equation}
\left\{
\begin{aligned}
\Vect{x}^{k+1} &= \argmin_{\Vect{x}}  \frac{1}{2} \left \Vert \Mat{A} \Vect{x} - \Vect{y} \right \Vert^2_2 + \frac{\rho}{2} \left \Vert \Mat{D} \Vect{x} - \Vect{z} + \Vect{u} \right \Vert^2_2 \\
\Vect{z}^{k+1} &= \argmin_{\Vect{z}} \mu \left\Vert \Mat{D} \Vect{x} - \Vect{z} \right\Vert_1+ \frac{\rho}{2} \left \Vert \Mat{D} \Vect{x} - \Vect{z} + \Vect{u} \right \Vert^2_2\\
\Vect{u}^{k+1} &= \Vect{u}^{k} + \Mat{D} \Vect{x}^{k+1} - \Vect{z}^{k+1},
\end{aligned}
\right.
\end{equation}
where $\Vect{u}$ is the scaled dual variable. The $\Vect{x}$-problem can be efficiently solved in the Fourier domain. The analytical solution is
\begin{equation}
\Vect{x}^{k+1} = \left( \Mat{A}^T \Mat{A} + \rho \Mat{D}^T \Mat{D} \right)^{-1} \left( \Mat{A}^T \Mat{A} + \rho \Mat{D}^T \Mat{D} \right).
\end{equation}
The $\Vect{z}$-problem has a closed-form solution, 
\begin{equation}
\Vect{z}^{k+1} = \mathcal{S}_{\mu / \rho} \left( \Mat{D} \Vect{x}^{k+1} + \Vect{u}^{k} \right),
\end{equation}
where $\mathcal{S}_{\kappa} \left( v \right) = \left( 1 - \kappa / \left| v \right| \right)_{+} v $ is an element-wise soft thresholding operator.
Finally $\Vect{u}$ is updated with the new $\Vect{x}$ and $\Vect{z}$.

We show more comparison results in Fig.~\ref{fig:supp_simulation} for the three candidate designs, the diffuser PSF~\cite{antipa2018diffusercam}, the Perlin PSF~\cite{boominathan2020phlatcam}, and our Voronoi-Fresnel PSF. The top two rows are from dataset~\cite{CAVE_0293} with lab setup scenes, and the bottom two rows are from the dataset~\cite{li2021multispectral} with natural indoor scenes. Our design outperforms the other two in both PSNR and SSIM.

\section{Comparison with auto-correlation}
The auto-correlation of the PSF is a concept related to MTF, and could in principle be used for optimization of optical phase elements, such as in Miniscope3D~\cite{yanny2020miniscope3d}. Since auto-correlation is not a single-number, it cannot be used directly as an optimization metric. A diffraction limited MTF must be taken as the reference. A major difference in the proposed MTFv concept is that, spatial and spectral information in the PSF are distilled into a single number that can be used for numerical optimization. Our MTFv metric requires no reference value, and evaluates the amount of useful information by itself.

\begin{figure*}[h!]
    \centering
    \includegraphics[width=\textwidth]{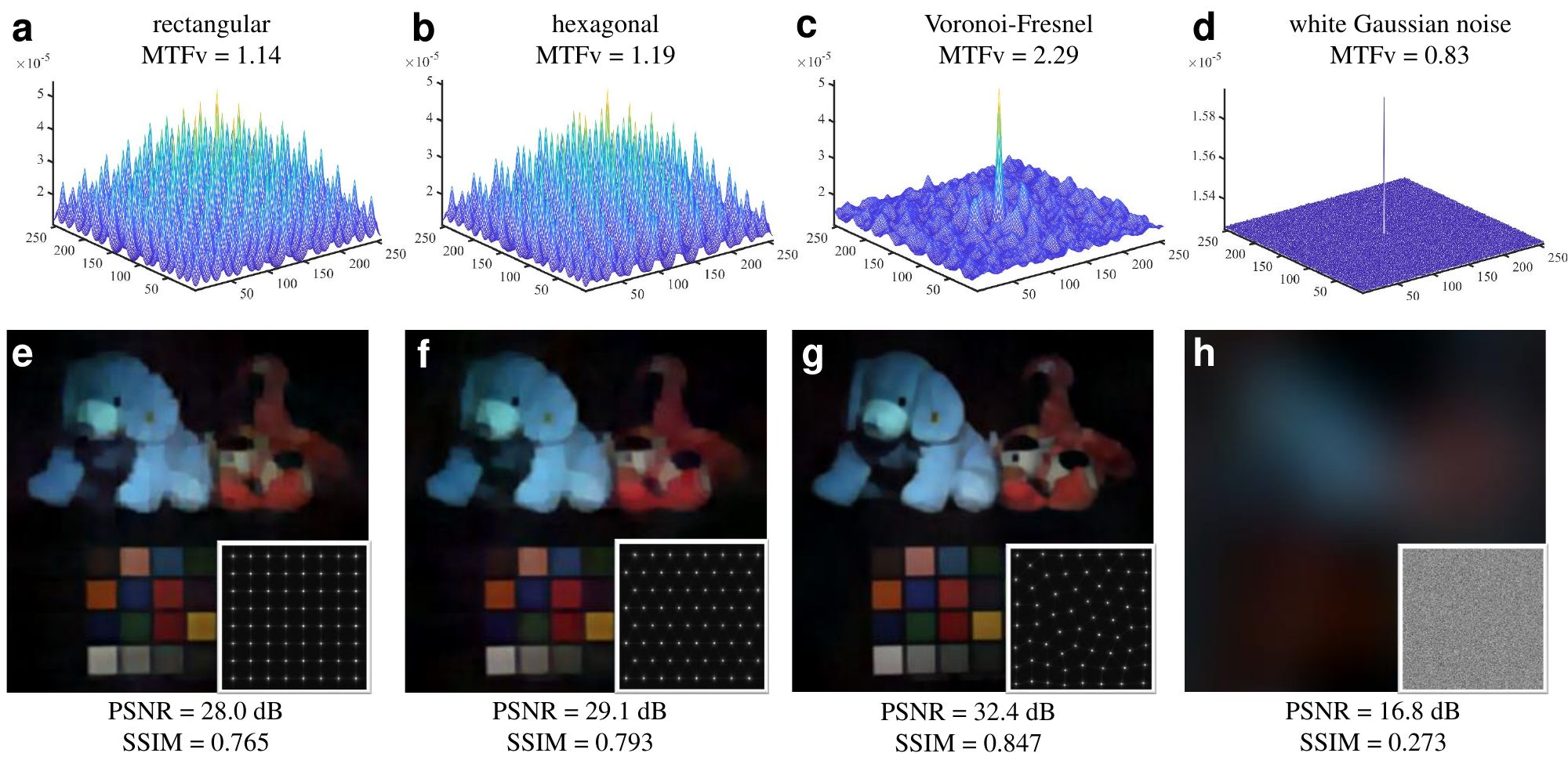}
    \caption{Image performance comparison with the auto-correlation metric. From (a) to (d) are the auto-correlation functions of the rectangular-grid PSF, hexagonal-grid PSF, optimized Voronoi-Fresnel PSF, and a white Gaussian noise PSF. From (e) to (h) are the reconstructed images for the PSFs shown above. The image quality is evaluated by PSNR and SSIM. The respective PSFs are shown in the bottom right corner insets.}
    \label{fig:xcorr}
\end{figure*}

In addition, we present an example to show that, auto-correlation may not be consistent in certain cases from the perspective of image quality, whereas our MTFv metric is more directly related to the final performance. Here we use four PSFs that show different auto-correlation properties. The first three are those we use in \tocheck{Fig.~4} in the main paper. Their auto-correlation functions are shown in Fig.~\ref{fig:xcorr}a-c. The fourth one is a white Gaussian noise pattern, which has an extremely sharp Dirac-like auto-correlation function (Fig.~\ref{fig:xcorr}d). As a comparison, the rectangular-grid and hexagonal-grid PSFs exhibit very large support, whereas the optimized Voronoi-Fresnel PSF has a moderate shape. We also compute their corresponding MTFv values, 1.14 (rectangular), 1.19 (hexagonal), 2.29 (Voronoi-Fresnel), and 0.83 (Gaussian). To evaluate the final image quality for these PSFs, we reconstruct the sharp images using the same parameters. The reconstructed images are shown in Fig.~\ref{fig:xcorr}e-h, with the PSNR and SSIM values shown below. Our Voronoi-Fresnel PSF yields the best PSNR (32.4~dB) and SSIM (0.765), significantly better than the other three.

This example indicates that, although the white Gaussian noise pattern has a strong peak in the auto-correlation, it fails in reconstructing a reasonable image. The rectangular and hexagonal PSFs show very large support in the auto-correlation, but not as good as the optimized Voronoi-Fresnel PSF. The result also emphasizes the importance of a proper reference quantity for the success of the auto-correlation metric. As a comparison, the proposed MTFv metric is consistently related with the image quality, and naturally rules out the white Gaussian noise PSF, so MTFv is a more robust metric for this design problem.

\section{Fabrication}
The experimental samples are fabricated by a combination of photolithography and reactive-ion etching (RIE) techniques. The substrate is a 4 inch fused silica wafer with a thickness of 0.5~mm. We binarize the optimized Voronoi-Fresnel phase profile into $2^4 = 16$ levels, and repeat 4 iterations of the basic photolithography with different masks and then RIE with doubled etching time. The masks are fabricated on soda lime substrates by laser direct writing on a Heidelberg $\mu$PG~501 mask maker. In each iteration, the wafer is first cleaned in Piranha solution at $115^{\circ}$C for 10 min, and dried with N$_2$. A 150-nm-thick Chromium (Cr) layer is deposited by sputtering on one side of the wafer. A 0.6-$\mu m$-thick layer of photoresist AZ1505 is then spin-coated on top of Cr after HMDS (Hexamethyldisilazane) vapor priming. To transfer the mask patterns to the wafer, we align the wafer with the mask on a contact aligner EVG6200~$\infty$ in the hard+vacuum mode, and then apply UV exposure with a dose of 9~mJ/cm$^2$. The photoresist is developed in AZ726MIF developer (2.38\% TMAH in H$_2$O) for 20~sec before De-Ionized water clean and N$_2$ drying. To transfer the patterns from the photoresist to the Cr layer, we wet-etch the wafer with Cr etchant (mixtures of HClO$_4$ and (NH$_4$)$_2$[Ce(NO$_3$)$_6$]) for 1 min and remove the residual photoresist with Acetone. In the RIE step, SiO$_2$ in the wafer is etched by plasma of 15 sccm CHF$_3$ and 5~sccm O$_2$ at 10~$^{\circ}$C. The etching depths are 75~nm, 150~nm, 300~nm and 600~nm respectively for a design wavelength at 550~nm. An additional Cr layer is deposited and etched to serve as the aperture to preserve the shift-invariance of PSFs.

\section{Prototype results}
\begin{figure*}[h!]
    \centering
    \includegraphics[width=\textwidth]{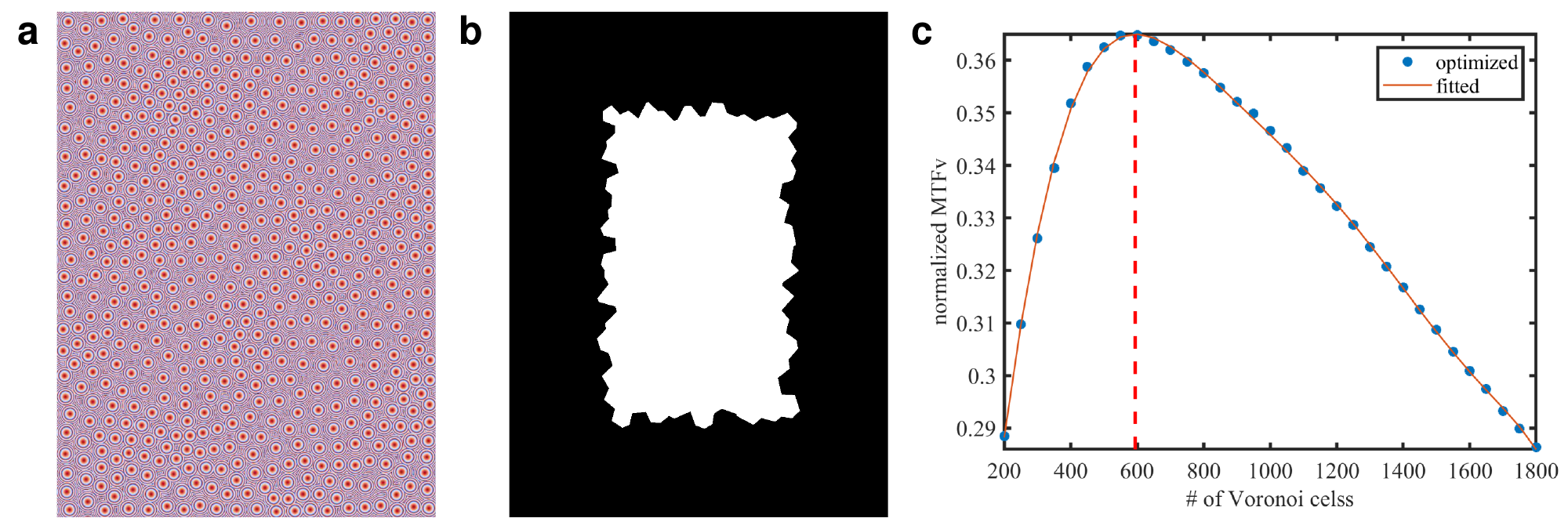}
    \caption{Design of the prototype Voronoi-Fresnel lensless camera. (a) Optimized full-area phase profile. (b) An aperture to maintain the PSF shift-invariance within $\pm \mathrm{20}^{\circ}$ field-of-view. (c) Best number of Voronoi-Fresnel cells in the parameter sweep step. The optimal number of cells is 594 without the aperture. Effective cells after applying the aperture is 424.}
    \label{fig:supp_prototype}
\end{figure*}

The prototype Voronoi-Fresnel phase is designed at 550~nm for $2\pi$ modulation. Considering the fabrication resolution and sensor pixel size, we fix the upsampling ratio of 3$\times$ for the optical element, i.e., 1.15~$\mu$m, which is well controlled by our fabrication method. The sensor has 1440 $\times$ 1080 pixels, and the Voronoi-Fresnel phase has 4320 $\times$ 3240 pixels. The optimized full-area phase profile is shown in Fig.~\ref{fig:supp_prototype}a. Our sensor has a field-of-view of $\pm \mathrm{20}^{\circ}$ in the horizontal direction, and $\pm \mathrm{15}^{\circ}$ in the vertical direction, so we design an aperture (Fig.~\ref{fig:supp_prototype}b) that excludes the cells outside of the field-of-view. The optimal number of Voronoi-Fresnel cells from the optimization is 594, as shown in Fig.~\ref{fig:supp_prototype}c after the parameter sweep step. The effective cells within the field-of-view is 424, which is about 71.4\% of the total number.

The final presentation of the image requires some necessary pre-processing and post-processing for better image presentation. We first demosaic the raw sensor data into color image data according to the sensor Bayer layout. Before image reconstruction, we normalize the blurred data by its norm in each color channel. For all the experimental results, the weights are all set to $\mu = 1e^{-7}$, and $\rho = 1e^{-5}$. In the post-processing, we use a simple gray world algorithm in MATLAB (\texttt{chromadapt}) for automatic white balancing in the linear color space. The illuminant is estimated by excluding 10 percentile of pixels. Finally gamma correction ($\gamma$ = 1.25) is applied. 

\begin{figure*}[h!]
    \centering
    \includegraphics[width=\textwidth]{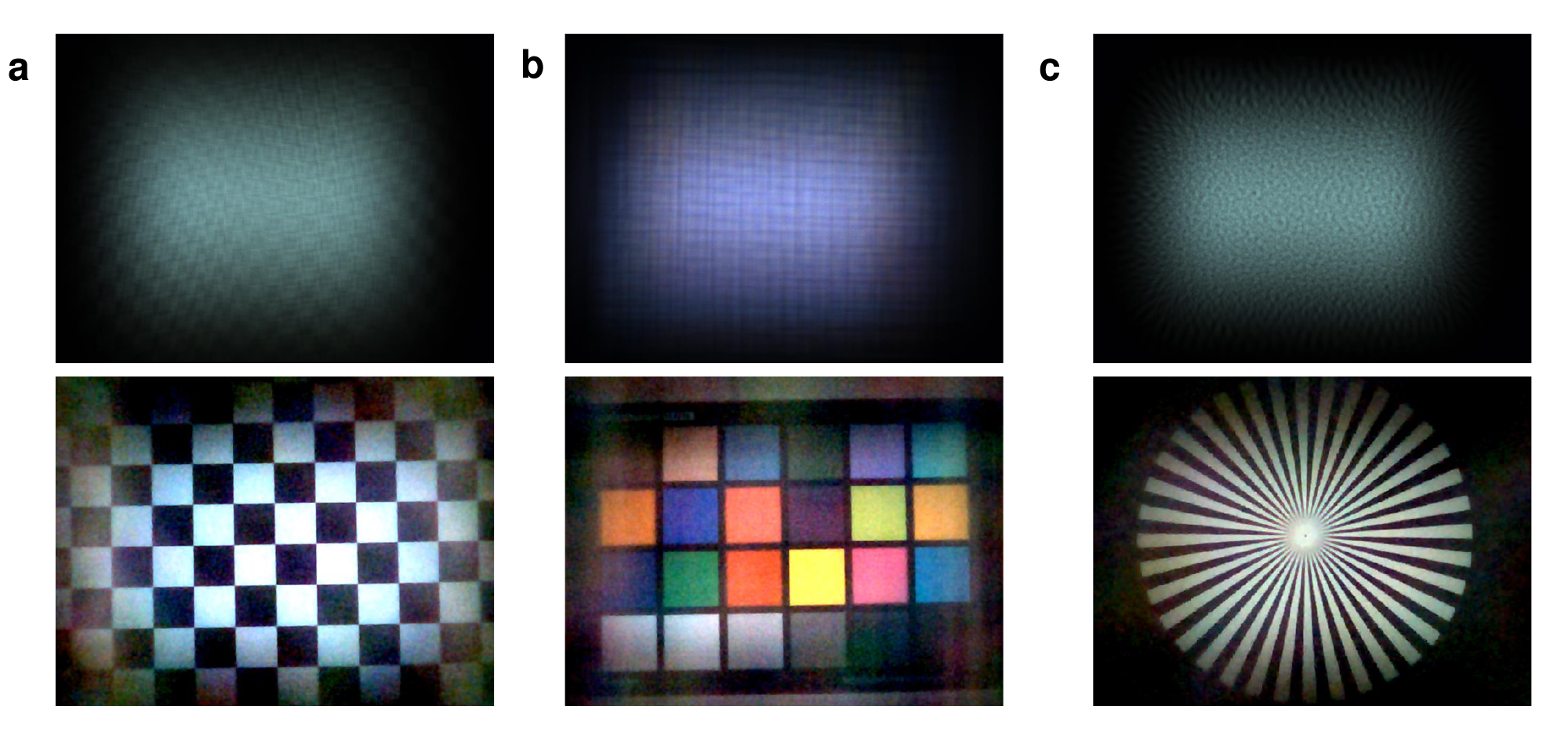}
    \caption{Prototype characteristic test results. The reconstruction of a checkerboard image in (a) shows little geometry distortion. The color checker in (b) indicates good color recovery, although residual color artifacts exist in the image border. (c) A Siemens star image shows uniform resolution preservation in all directions.}
    \label{fig:supp_prototype_geometry}
\end{figure*}

We present additional characteristic test results for the prototype in Fig.~\ref{fig:supp_prototype_geometry}. First, we evaluate the geometry distortion using a checkerboard target. As shown in Fig.~\ref{fig:supp_prototype_geometry}a, the geometry is restored very well across the $\pm \mathrm{20}^{\circ} \times \pm \mathrm{15}^{\circ}$ field-of-view. On the border regions, there are residual chromatic artifacts, however. This may arise from two factors. One is the off-axis aberrations of the base Fresnel phase, and the other is the difference in PSFs from on-axis to off-axis leading to the drop in reconstruction quality. Second, we evaluate the color fidelity using a color checker target in Fig.~\ref{fig:supp_prototype_geometry}b. Despite the residual color artifacts in the border regions, the overall color fidelity is retrieved from the raw data. Potential improvement could be a more advanced white balancing algorithm other than the simple gray world algorithm used here. Last, we evaluate the spatial resolution variation using a Siemens star target in Fig.~\ref{fig:supp_prototype_geometry}c. A uniform spatial resolution change is observed from the reconstructed image, demonstrating again that the MTF optimization is effective with uniform response.

\begin{figure*}[h!]
    \centering
    \includegraphics[width=0.9\textwidth]{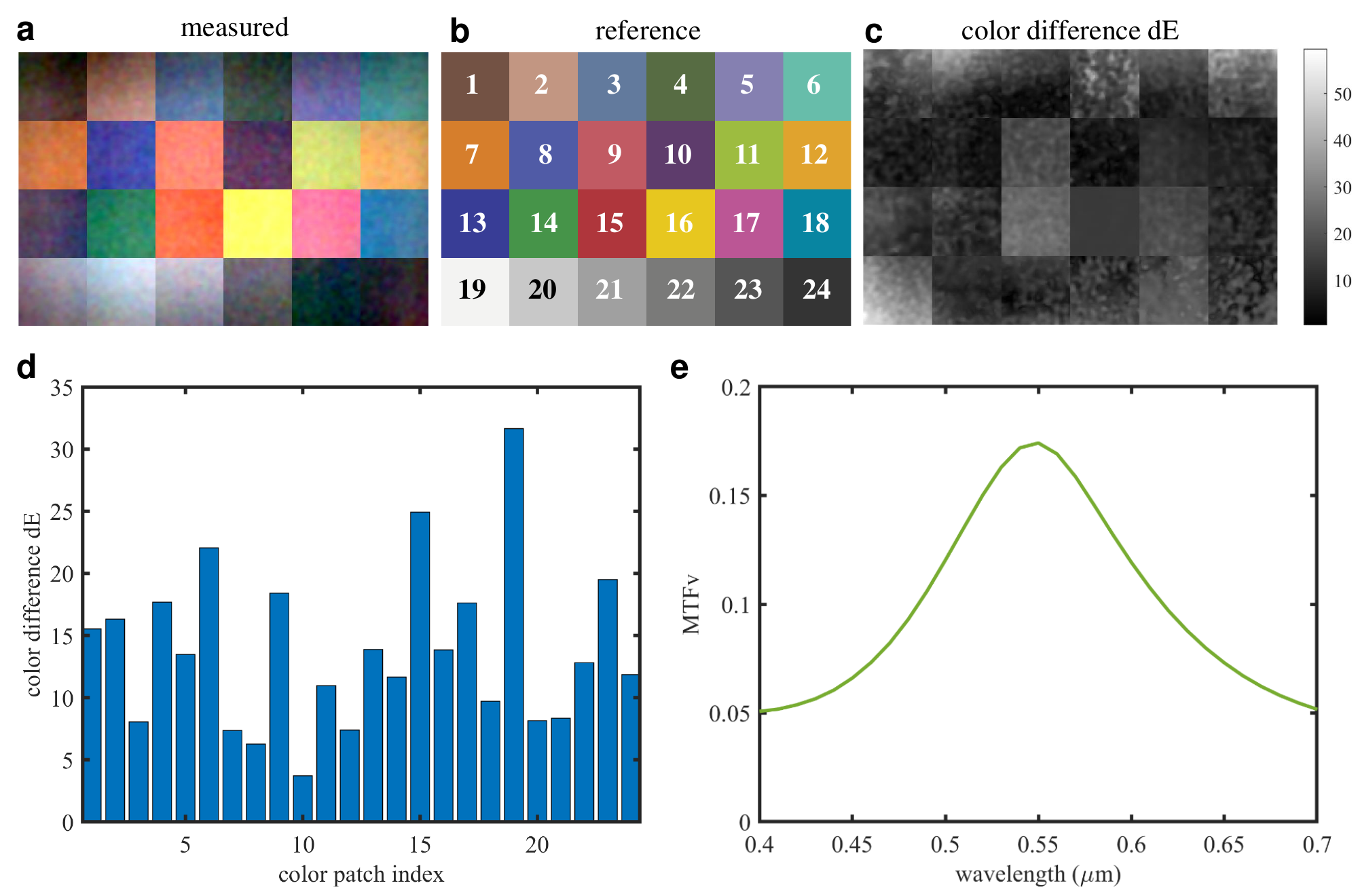}
    \caption{Color difference analysis. (a) Extracted color patches from the reconstructed color checker image. (b) Reference color patches with true RGB values. Indices are labeled for each patch. (c) Color different $dE$ map calculated from (a) and (b) using the CIEDE2000 standard. (d) Color difference $dE$ values calculated with averaged RGB values in each color patch. (e) MTFv varies with wavelength in the visible band for the Voronoi-Fresnel phase used in the prototype.}
    \label{fig:supp_color_fidelity}
\end{figure*}

Color reproduction remains a challenge in the current prototype. To analyze the color fidelity, we extract the color patches from the reconstructed image of the color checker (Fig.~\ref{fig:supp_prototype_geometry}a), and tile them side by side according to their original orders in the color checker to synthesize an image in Fig.~\ref{fig:supp_color_fidelity}a. Each extracted patch has 50 $\times$ 50 pixels. As a reference, we create a reference color patch image from the corresponding true RGB values with the same size, as shown in Fig.~\ref{fig:supp_color_fidelity}b, with their indices labeled. The color fidelity is measured as the color difference $dE$ using the CIEDE2000 standard~\cite{sharma2005ciede2000}. We can visualize the pixel-wise color difference with the $dE$ map in Fig.~\ref{fig:supp_color_fidelity}c. Since the $dE$ map varies within each color patch due to the noisy reconstruction results, we take the average value in each color patch as the reconstructed color values. The color difference $dE$ values are then calculated and plotted in Fig.~\ref{fig:supp_color_fidelity}d. The maximum $dE$ (largest color difference) is 31.6 at index 19, which is the ``White'' patch, and the the minimum $dE$ (smallest color difference) is 3.7 at index 10, which is the ``Purple'' patch. This also agrees with the visual perception.

We attribute the quality of color fidelity in our prototype mainly to the residual chromatic aberrations, and the simple white balancing algorithm we currently use. Since the base Fresnel phase is static at one single wavelength (550~nm in our prototype), the PSF geometry changes if the illumination wavelength is different, and hence the MTFv is also a function of wavelength. We optimize for the spectral integral of the MTF, but not the cross differences between wavelengths. We evaluate this chromatic effect by plotting the MTFv with respect to wavelength across the visible band from 400~nm to 700~nm, as shown in Fig.~\ref{fig:supp_color_fidelity}e. The MTFv drops around 71\% for both the short and long ends of the wavelength range, compared with the design wavelength at 550~nm. This could be mitigated by using a base phase function that is optimized achromatic, instead of the static Fresnel phase we currently use. Further improvement can be made to use more advanced white balance algorithms to improve the color fidelity.

\change{Although we use the panchromatic PSF in the visible band in our design, there are still residual chromatic aberrations in the current result. We attribute mainly two important factors for the residual chromatic aberrations. First, the fixed base Fresnel phase in each cell is inherently dispersive, which is a fundamental property of all the diffractive optical elements. It could be possible to use an optimized achromatic base phase in each cell, which would require additional efforts. Second, the simple image reconstruction algorithm we adopt here does not account for chromatic aberration correction. The total variation regularization only takes care of spatial structures in the image. It would then be more advantageous to employ neural networks for the reconstruction to alleviate chromatic aberrations.}

\begin{figure*}[h!]
    \centering
    \includegraphics[width=0.9\textwidth]{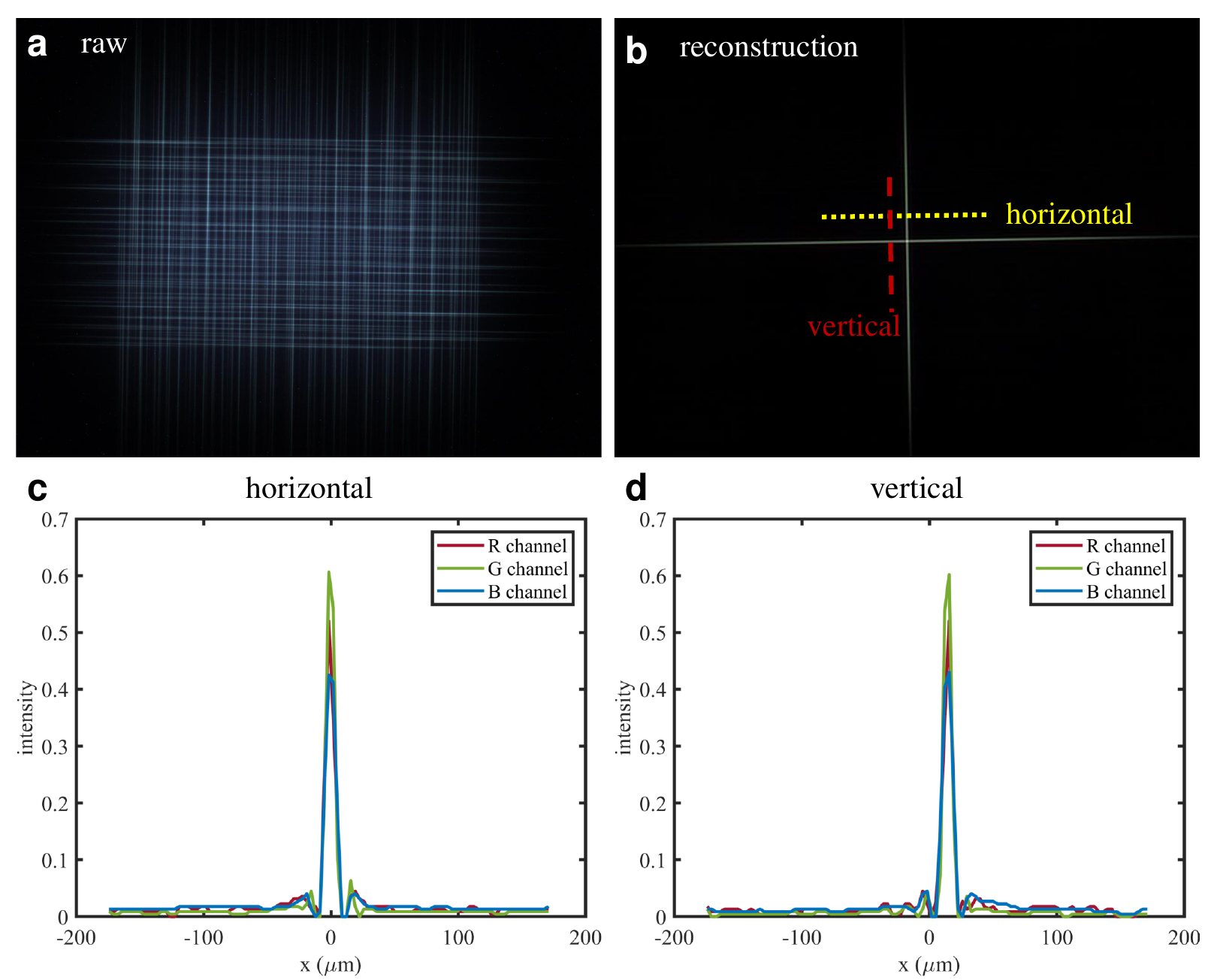}
    \caption{Resolution measurement. (a) Raw capture of the cross-hair target. (b) Reconstructed image of the cross-hair target. The yellow dotted line indicates where the horizontal cross-section is taken, and the red dash line indicates where the vertical cross-section is taken. (c) Horizontal cross-section of (b). (d) Vertical cross-section of (b). The spot diameter is measured around the lines as the intensity first reaches zero.}
    \label{fig:supp_resolution}
\end{figure*}

To evaluate the resolution of the prototype, we capture a cross-hair target, and reconstruct the final image. The raw data is shown in Fig.~\ref{fig:supp_resolution}a, and the reconstructed image is shown in Fig.~\ref{fig:supp_resolution}b. We can plot the cross-sections in the horizontal and vertical directions. The spot diameter is measured around the lines as the intensity first reaches zero. Since we know the pixel size, we can calculate the physical diameters. In the horizontal direction, the diameters are measured as 17.25~$\mu$m, 17.25~$\mu$m, 20.7~$\mu$m for the red, green and blue channels respectively. The average diameter is 18.4~$\mu$m. In the vertical direction, the diameters are 20.7~$\mu$m, 17.25~$\mu$m, 24.15~$\mu$m for the red, green and blue channels respectively. The average diameter is 20.7~$\mu$m. To compare with the theoretical value, we calculate the effective diameter of all the Voronoi-Fresnel cells. The ideal spot diameter is 15.7~$\mu$m, so the resolution of the prototype is indeed close to the theoretical value.

\begin{figure*}[h!]
    \centering
    \includegraphics[width=\textwidth]{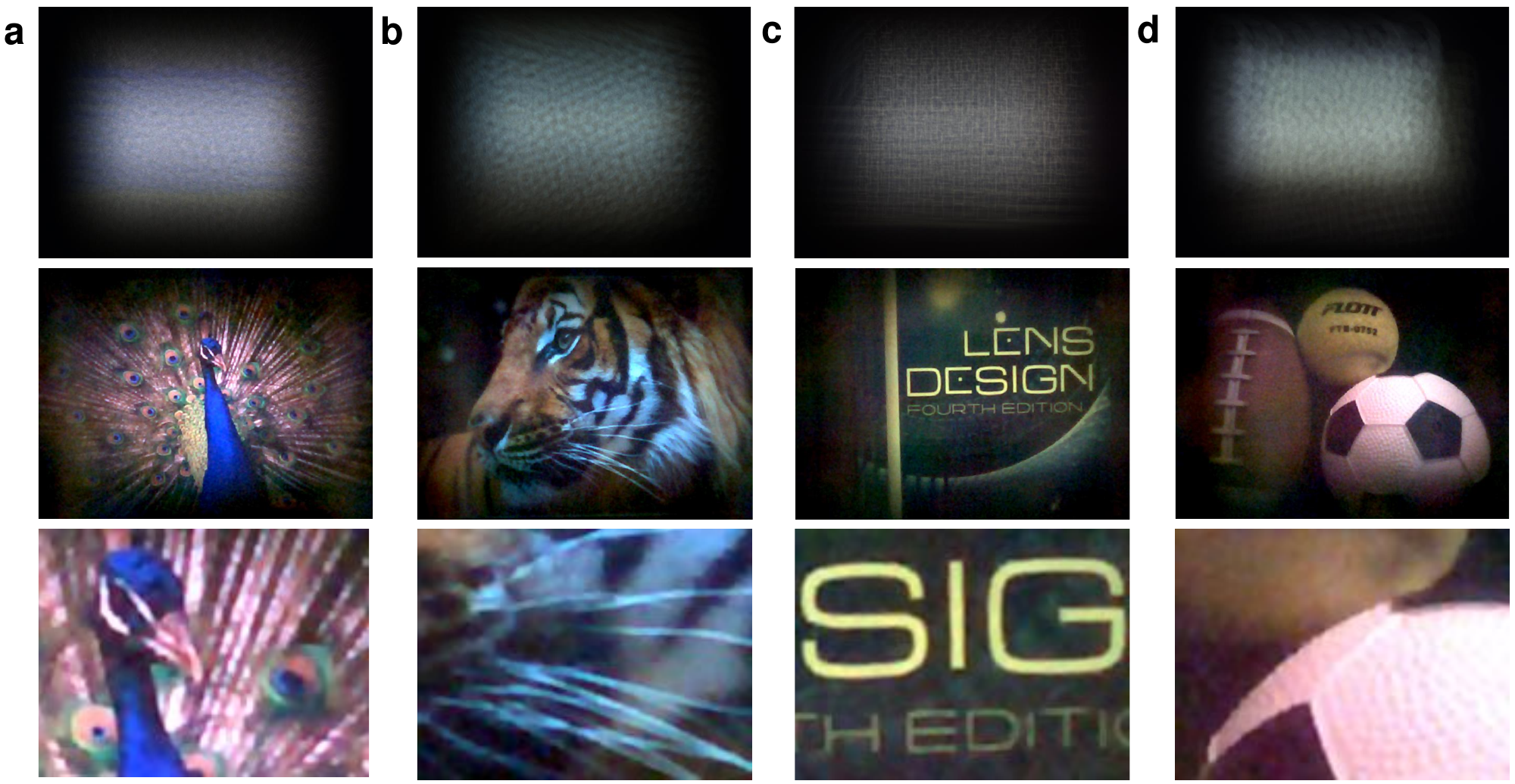}
    \caption{Additional prototype results. (a) and (b) show results for self-illuminating images displayed on a computer monitor. (c) and (d) show results for real objects with ambient illumination. Top row are the captured raw data; middle row are reconstructed images; and bottom row are zoom-in details.}
    \label{fig:supp_prototype_additional}
\end{figure*}

Additionally, we present more example results in Fig.~\ref{fig:supp_prototype_additional}. Again, the results in Fig.~\ref{fig:supp_prototype_additional}a and \ref{fig:supp_prototype_additional}b are captured from self-illuminating images displayed on a monitor, and Fig.~\ref{fig:supp_prototype_additional}c-d are real objects with ambient illumination. Details in the objects are well preserved in both cases.

